\pdfoutput=1
\documentclass[11pt]{article}

\usepackage{acl}

\usepackage{times}
\usepackage{latexsym}

\usepackage[T1]{fontenc}
\usepackage[utf8]{inputenc}

\usepackage{microtype}

\usepackage{inconsolata}

\usepackage{amssymb}% http://ctan.org/pkg/amssymb
\usepackage{pifont}% http://ctan.org/pkg/pifont
\usepackage{siunitx}
	\DeclareSIUnit{\quantity}{\relax}
	\DeclareSIUnit{\words}{words}
	\DeclareSIUnit{\sentences}{sentences}

\usepackage{graphicx}
\usepackage{graphicx}
\usepackage{multirow}
\usepackage{array}
\usepackage{multicol}
\usepackage{tablefootnote}
\usepackage{booktabs}
\usepackage{hyperref}
\usepackage[normalem]{ulem}

\newcommand{\cmark}{{\color{teal} \ding{51}}}%
\newcommand{\xmark}{{\color{red} \ding{55}}}%

\newcommand{\github}{\url{https://github.com/JHU-CLSP/Kreyol-MT}}

\title{Krey\`{o}l-MT: Building MT for Latin American, Caribbean and Colonial African Creole Languages}

\author{
Nathaniel R. Robinson$^{1}$ 
\quad Raj Dabre$^{3}$ 
\quad Ammon Shurtz$^{2}$
\quad Rasul Dent$^{4}$ \\
\quad \textbf{Onenamiyi Onesi}$^{5}$
\quad \textbf{Claire Bizon Monroc}$^{4}$
\quad \textbf{Lo\"{i}c Grobol}$^{6}$
\quad \textbf{Hasan Muhammad}$^{1}$ \\
\quad \textbf{Ashi Garg}$^{1}$
\quad \textbf{Naome A. Etori}$^{7}$
\quad \textbf{Vijay Murari Tiyyala}$^{1}$
\quad \textbf{Olanrewaju Samuel}$^{8}$ \\
\quad \textbf{Matthew Dean Stutzman}$^{2}$
\quad \textbf{Bismarck Bamfo Odoom}$^{1}$
\quad \textbf{Sanjeev Khudanpur}$^{1}$ \\
\quad \textbf{Stephen D. Richardson}$^{2}$
\quad \textbf{Kenton Murray}$^{1}$
\\
$^1$Johns Hopkins University, USA; $^2$Brigham Young University, USA;\\
$^3$National Institute of Information and Communications Technology, Japan;\\ 
$^4$Inria Paris; $^5$Nile University of Nigeria; $^6$Universit\'{e} Paris Nanterre;\\ 
$^7$University of Minnesota - Twin Cities, USA; $^8$University of Toronto\\
 \texttt{nrobin38@jhu.edu}\\
}

\begin{document}
\maketitle

\begin{abstract}
A majority of language technologies are tailored for a small number of high-resource languages, while relatively many low-resource languages are neglected. 
One such group, Creole languages, have long been marginalized in academic study, though their speakers could benefit from machine translation (MT). 
These languages are predominantly used in much of Latin America, Africa and the Caribbean. 
We present the largest cumulative dataset to date for Creole language MT, including \qty{14.5}{\mega\quantity} unique Creole sentences with parallel translations---\qty{11.6}{\mega\quantity} of which we release publicly, and the largest bitexts gathered to date for \num{41} languages---the first ever for \num{21}. 
In addition, we provide MT models supporting all \num{41} Creole languages in \num{172} translation directions. 
Given our diverse dataset, we produce a model for Creole language MT exposed to more genre diversity than ever before, which outperforms a genre-specific Creole MT model on its own benchmark for \num{26} of \num{34} translation directions. 
\end{abstract}

\section{Motivation}
\label{sxn:intro}

\begin{figure}[th]
    \centering
    \includegraphics[width=.85\linewidth]{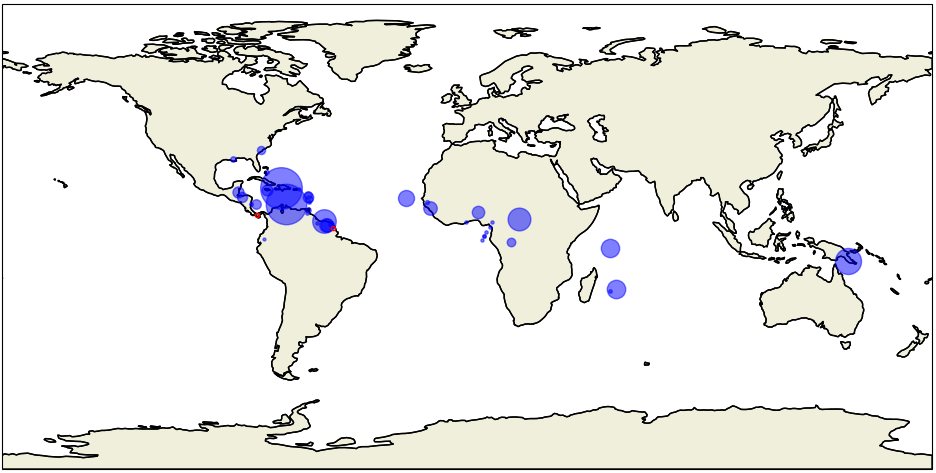}
    \includegraphics[width=.85\linewidth]{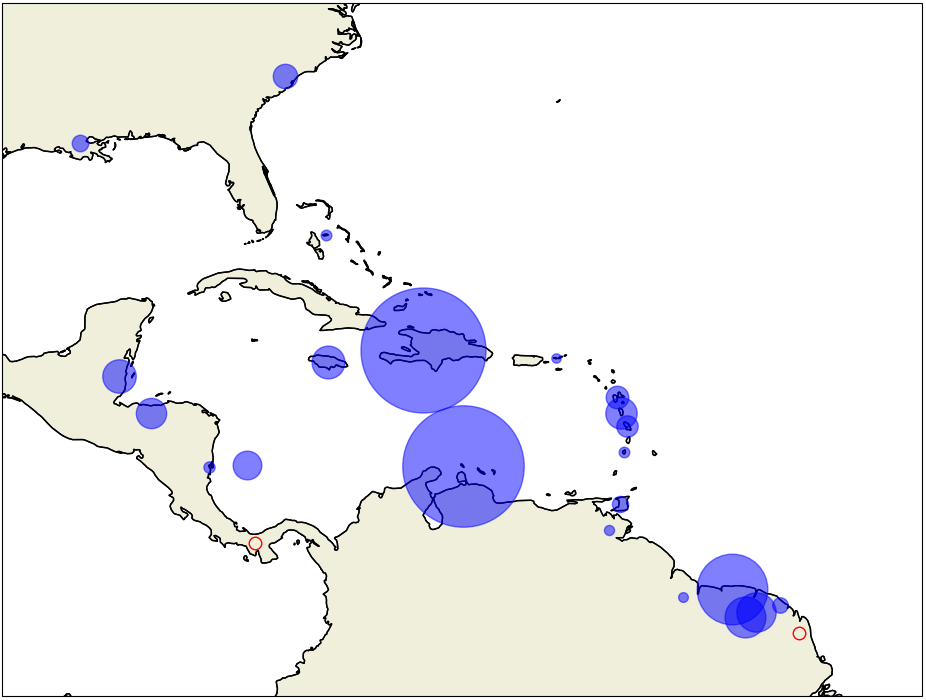}
    \caption{Dataset sizes plotted geographically, with centroids from Glottolog. Each circle's area is proportionate to the square root of the data amount for each language, to facilitate viewing.} 
    \label{fig:maps}
\end{figure}

From northern Brazil to the Gulf of Mexico, spanning an area including the Caribbean and Central American west coast, lies the Creole "civilizational region" \cite{glissant2008creolization}. 
One of its chief characteristics: a multiplicity of Creole languages, born from contact of African language speakers with European languages in the colonial era. 
% \nrr{List any facts and stats about Creole languages in Latin America?} \nrr{David}
Low-resource Creole languages are widely spoken here and throughout the world \citep{rickford2017language, Mufwene2008-il, bartens2021making, Velupillai2015-tz}.  
Historic linguistic marginalization has stymied their technological advancement: few language technologies exist for these languages despite their many speakers \citep{lent2023creoleval}. 

Better MT could greatly benefit Creole language speakers. 
Many live in areas where their language is in the minority. 
Panama and Costa Rica are home to communities of West Indian descent who have maintained Creole languages \citep{conniff1983black, herzfeld1980limon}. 
Large Haitian-speaking communities live in the Dominican Republic \cite{zhong2023haiti}, Chile, Mexico \citep{audebert2017recent}, Brazil \cite{terry2019new}, and the Bahamas \citep{perry2023real, mccartney2013rise, knowles2018case}.
Language is one of the first obstacles to immigrants' social integration, and many report daily reliance on MT \cite{neto2020latin}. 
The lack or low accuracy of such technologies can thus contribute to social exclusion.

Multiple Creole-speaking communities regularly fall victim to natural disasters \cite{heinzelman2010crowdsourcing, margesson2010haiti, rasmussen2015development,look2019resilience}. 
As the frequency of Atlantic hurricanes may be accelerated by global climate change \cite{hosseini2018influence}, machine translation can provide useful tools to facilitate communication during international relief efforts \citep{lewis2010haitian, hunt2019ethics}.

Yet colonial-era stigmas dismissing Creole languages as broken or incomplete persist, and serve as justifications to advantage European languages at their expense \citep{alleyne1971acculturation,degraff2003against}. 
Association with lower economic status and limited use in official settings then inhibit data collection and Natural Language Processing (NLP) development for these languages \citep{lent2023creoleval}. 
We expand on \citet{lent2023creoleval,lent-etal-2022-creole} to unify community efforts in advancing Creole NLP. 

In addition to meeting community needs, Creole language MT presents avenues of exploration for low-resource NLP. 
Many Creole languages have documented linguistic relationships with high-resource languages, as well as lexical and morphosyntactic proximity to each other \citep{rickford2017language}. 
(See \S\,\ref{sxn:meth}.) 
As such, they have potential for cross-lingual transfer \cite{lent2023creoleval}, a powerful technique for low-resource NLP \cite{pfeiffer2020mad,kim-etal-2019-effective}. 
This potential presents an opportunity to develop technologies for many Creole languages at once. 
But since state-of-the-art NLP methods rely on machine learning, this development is not possible without data.
We present an expansive dataset for MT of Creole languages, as a meaningful first step towards developing their technologies. We contribute:

\begin{itemize}
    \item The largest, most genre-diverse MT datasets ever compiled for 41 Creole languages, including the first ever for 21 Creole languages
    \item A public dataset of 11.6M aligned sentences and 3.4M monolingual sentences for 40 Creole languages\footnote{Visit our repository \github{} for software, models, and data download.}
    \item Public models achieving state-of-the-art performance on a published Creole language benchmark for 26 language directions
\end{itemize}

\section{Background and Related Work}
\label{sxn:rw}

Despite its potential benefits, previous Creole language MT development has been scarce. 
Google Translate,\footnote{\url{https://translate.google.com}} 
a common MT interface, only supports one Latin American Creole language (Haitian). 
NLLB-200 \citep{team2022language}, a state-of-the-art MT model in the number of languages it supports (204), only supports five Creole languages. 
The emergence of large language models like ChatGPT,\footnote{\url{chat.openai.com}} trained on massive unlabeled datasets, presents encouraging potential for more universal MT support. 
However, recent studies indicate that LLMs are unable to compete with supervised MT for low-resource languages \citep{robinson-etal-2023-chatgpt,zhu2023multilingual}. 
Researchers have also speculated that LLMs’ MT capabilities in high-resource languages are due to curated bitexts in their training data \citep{briakou-etal-2022-bitextedit}. 
Hence, developing bitext corpora for low-resource languages is still important in building MT for them. 

Some prior works have approached Creole language MT, like the inclusion of early Haitian-English MT in the DIPLOMAT system \citep{frederking1997diplomat} and the 2011 Workshop on Statistical Machine Translation (WMT) \citep{callison2011findings}. Creole languages that are the focus of more recent MT research include Haitian and Jamaican \citep{robinson-etal-2022-data,robinson2023african}, Naija \citep{adelani-etal-2022-thousand,ogueji2019pidginunmt}, Mauritian \citep{dabre2022kreolmorisienmt}, Sranan Tongo \citep{zwennicker2022towards}, and Singlish \citep{wang2017universal,liu2022singlish}. 
Additional NLP research has been conducted in sentiment analysis and named entity recognition for Naija \citep{oyewusi2020semantic,muhammad2022naijasenti,muhammad-etal-2023-semeval,adelani2021masakhaner}; syntactic analysis for Singlish \citep{wang2017universal}, Naija \citep{caron2019surface}, and Martinican \citep{mompelat2022parse}; and inference for Jamaican \citep{armstrong2022jampatoisnli}. 

These works, while valuable steps forward, have been limited in their scope of languages. 
The prior work most comparable to ours is the recent CreoleVal \citep{lent2023creoleval}, which focused on building datasets for 28 Creole languages. 
They focused on machine comprehension, relation classification, and MT. 
\citeauthor{lent2023creoleval}'s (\citeyear{lent2023creoleval}) MT dataset is a significant contribution to our work. 
They do not publicly release their MT datasets, however. 
While portions of the data we collected (including part of the CreoleVal set) are not publicly releasable, we release 11.6M aligned bitext sentences and 3.4M monolingual sentences in 40 languages.
The genre coverage of CreoleVal's solely religious and Bible text data is another limitation shared by prior works, which can preclude general-purpose MT.
For example, \citeauthor{robinson2023african}'s (\citeyear{robinson2023african}) Haitian model trained on primarily religious texts achieves an impressive BLEU score of 68.0 on its same-genre test set, but a score of 14.7 on a Wikipedia-style test set. 
This indicates that Creole language MT models trained on specific genres may not be applicable to general domains. 
We provide the most diverse Creole language MT data yet, both in terms of languages and genres. (See \S\,\ref{sxn:genres}.)

Other multilingual MT works include some Creole languages. 
In addition to \citeauthor{team2022language}’s (\citeyear{team2022language}) five included Creole languages,  \citeauthor{yuan2023lego}’s (\citeyear{yuan2023lego}) multilingual Lego-MT model supports 8 Creole languages (out of 433 total). 
Our work is a significant expansion on this, making a meaningful contribution to the broader effort of low-resource dataset curation---including the WMT'23 shared task on the topic \cite{sloto2023findings}, which our work follows by involving noisy data filtering \cite{minh2023fast} and document alignment \cite{steingrimsson2023sentence}.

\section{Methodology and Dataset}
\label{sxn:meth}

The languages we include in our study are in Table~\ref{tab:langs}. 
We retrieve each language’s vitality, official recognized status by a political entity (\textit{Off.}), and number of speakers from Ethnologue.\footnote{\url{https://www.ethnologue.com}}
We label a language as a majority language (\textit{Maj.}) if it is spoken by a majority of the population in one of its native countries or territories. 
This project includes languages from \num{24} Latin American and Caribbean countries and territories. (See Figure~\ref{fig:maps}.)

\begin{table*}
    \setlength{\tabcolsep}{3.5pt}
    \renewcommand{\arraystretch}{1.20}
    \centering
    \small
    \begin{tabular}{
        l % lang
        >{\ttfamily}{c} % ISO 639-3 code
        >{\ttfamily}{c} % Glottocode
        l % Native to
        r % Vitality
        S[table-format=6.1] % L1 speakers
        S[table-format=6.1] % L2 speakers
        c % Official status
        c % Majority language
    }
        \toprule
        & & & & & \multicolumn{2}{c}{\textbf{Speakers}~/~\unit{\kilo\quantity}} & &  \\
        \cmidrule(lr){6-7}
        \textbf{Language} & {\rmfamily\textbf{ISO}} & \rmfamily\textbf{Glottocode} & \textbf{Native to\dots} & \textbf{Vitality} & \textit{L1} & \textit{L2} & \textbf{Off.} & \textbf{Maj.} \\
        \midrule
        \uline{Saint Lucian Patois} & \textbf{acf} & sain1246 & Saint Lucia & Stable & 760 & {-} & \xmark & \cmark \\
        \uline{Bahamian Creole} & \textbf{bah} & baha1260 & Bahamas & Stable & 340 & {-} & \cmark & \cmark \\
        Berbice Dutch & \textbf{brc} & berb1259 & Guyana & Extinct & 0 & 0 & \xmark & \xmark \\
        \uline{Belizean Kriol}  & \textbf{bzj} & beli1260 & Belize & Institutional & 170 & {-} & \xmark & \cmark \\
        \uline{Miskito Coast Creole} & \textbf{bzk} & nica1252 & Nicaragua & Institutional & 18 & {-} & \xmark & \xmark \\
        \uline{Garifuna}    & \textbf{cab} & gari1256 & Central America & Endangered & 120 & {-} & \xmark & \xmark \\
        Negerhollands & \textbf{dcr} & nege1244 & U.S. Virgin Islands & Extinct & 0 & 0 & \xmark & \xmark \\
        \uline{Ndyuka}  & \textbf{djk} & ndyu1242 & Suriname, French Guiana & Stable & 68 & {-} & \xmark & \xmark \\
        \uline{Guadeloupean Creole} & \textbf{gcf} & \textbf{guad1243} & Guadeloupe & Stable & 580 & {-} & \xmark & \cmark \\
        \uline{Martinican Creole} & gcf & \textbf{mart1259} & Martinique & Stable & 520 & {-} & \xmark & \cmark \\
        \uline{French Guianese Creole} & \textbf{gcr} & guia1246 & French Guiana & Stable & 180 & {-} & \xmark & \cmark \\
        \uline{Gullah} & \textbf{gul} & gull1241 & South Carolina, Georgia & Endangered & 250 & {-} & \xmark & \xmark \\
        Creolese & \textbf{gyn} & creo1235 & Guyana & Stable & 720 & {-} & \cmark & \cmark \\
        Haitian & \textbf{hat} & hait1244 & Haiti & Institutional & 13000 & 69 & \cmark & \cmark \\
        \uline{San Andr\'{e}s-Providencia} & \textbf{icr} & sana1297 & Colombia & Stable & 12 & {-} & \xmark & \xmark \\
        \uline{Jamaican Patois} & \textbf{jam} & jama1262 & Jamaica & Stable & 3100 & 3.7 & \xmark & \cmark \\
        \uline{Karip\'{u}na} & \textbf{kmv} & kari1301 & Brazil & Endangered & 2.4 & {-} & \xmark & \xmark \\
        \uline{Louisiana Creole} & \textbf{lou} & loui1240 & Louisiana & Endangered & 4.8 & {-} & \xmark & \xmark \\
        Media Lengua & \textbf{mue} & medi1245 & Ecuador & Endangered & 2.6 & {-} & \xmark & \xmark \\
        \uline{Papiamento} & \textbf{pap} & papi1253 & Aruba, Cura\c{c}ao, Bonaire & Institutional & 350 & 20 & \cmark & \cmark \\
        \uline{San Miguel Creole} & \textbf{scf} & sanm1305 & Panama & Extinct & 0 & 0 & \xmark & \xmark \\
        \uline{Saramaccan} & \textbf{srm} & sara1340 & Suriname, French Guiana & Stable & 35 & {-} & \xmark & \xmark \\
        \uline{Sranan Tongo} & \textbf{srn} & sran1240 & Suriname & Institutional & 520 & 150 & \xmark & \cmark \\
        Vincentian Creole & \textbf{svc} & vinc1243 & Saint Vincent & Stable & 110 & {-} & \xmark & \cmark \\
        \uline{Trinidadian Creole} & \textbf{trf} & trin1276 & Trinidad & Stable & 1000 & {-} & \xmark & \cmark \\
        \midrule
        Angolar & \textbf{aoa} & ango1258 & S\~{a}o Tom\'{e} and Pr\'{i}ncipe & Endangered & 12 & {-} & \xmark & \xmark \\
        Saotomense & \textbf{cri} & saot1239 & S\~{a}o Tom\'{e} and Pr\'{i}ncipe & Endangered & 56 & {-} & \xmark & \xmark \\
        \uline{Seychellois Creole} & \textbf{crs} & sese1246 & Seychelles & Institutional & 88 & {-} & \cmark & \cmark \\
        Annobonese & \textbf{fab} & fada1250 & Equatorial Guinea & Stable & 6.6 & {-} & \xmark & \xmark \\
        Fanakalo & \textbf{fng} & fana1235 & South Africa & Endangered & 0 & {5.1} & \xmark & \xmark \\
        Pichi & \textbf{fpe} & fern1234 &  Equatorial Guinea & Institutional & 15 & 190 & \xmark & \xmark \\
        Ghanaian Pidgin & \textbf{gpe} & ghan1244 & Ghana & Institutional & 2.0 & 5000 & \xmark & \xmark \\ 
        \uline{Kabuverdianu} & \textbf{kea} & kabu1256 & Cape Verde & Institutional & 1200 & 14 & \xmark & \cmark \\
        Krio & \textbf{kri} & krio1253 & Sierra Leonne & Institutional & 820 & 7400 & \xmark & \cmark \\
        Kituba & \textbf{ktu} & kitu1246 & Central Africa & Institutional & 12000 & 800 & \cmark & \cmark \\
        \uline{Mauritian} & \textbf{mfe} & mori1278 & Mauritius & Institutional & 1000 & 6.5 & \xmark & \cmark \\
        \uline{Naija} & \textbf{pcm} & nige1257 & Nigeria & Institutional & 4700 & 120000 & \xmark & \cmark \\
        Guinea-Bissau Creole & \textbf{pov} & uppe1455 & Guinea-Bissau & Institutional & 340 & 1500 & \xmark & \cmark \\
        Principense & \textbf{pre} & prin1242 & S\~{a}o Tom\'{e} and Pr\'{i}ncipe & Endangered & 0.2 & {-} & \xmark & \xmark \\
        \uline{R\'{e}union Creole} & \textbf{rcf} & reun1238 & R\'{e}union & Stable & 810 & {-} & \xmark & \cmark \\
        Sango & \textbf{sag} & sang1328 & Central African Rep. & Institutional & 620 & 4600 & \cmark & \cmark \\
        \uline{Tok Pisin} & \textbf{tpi} & tokp1240 & Papua New Guinea & Institutional & 130 & 4000 & \cmark & \cmark \\
        Cameroonian Pidgin & \textbf{wes} & came1254 & Cameroon & Institutional & 12000 & {-} & \xmark & \cmark \\
        \bottomrule
    \end{tabular}
    \caption{Above are Creole languages of the Americas (Latin America, the Caribbean, and surrounding area). Below are Creole languages of Africa, as well as Tok Pisin. We refer to languages by their \textbf{bolded} language codes. \uline{Underlined} languages are those on which we focused for data gathering, outlined in \S\,\ref{sxn:meth}. The last two columns indicate respectively whether the language has official status and whether it is a majority language in one of its native countries/territories.
    }
    \label{tab:langs}
\end{table*}

As mentioned in \S\,\ref{sxn:rw}, \citeauthor{lent2023creoleval}'s (\citeyear{lent2023creoleval}) CreoleVal work is most comparable to our own. 
But the set of languages we include differs from theirs. 
Their focus was on Creole languages in general, while ours is narrowed to those of the Americas, allowing us to dive deeper and include more languages with greater linguistic commonalities. 
The languages we include have shared patterns of historical formation. 
They generally have morphosyntactic commonalities with Niger-Congo languages \citep{kouwenberg2004echoes, castillo2006emergence} and large-scale lexical overlap with Romance and Germanic languages  \citep{valdman2000evolution, winforf1997reexamining}.
For example, Haitian and Jamaican have extensively borrowed vocabulary from older forms of French and English, respectively, but they are morphosyntactically closer to Gbe, Kwa, and Igboid languages \citep{lefebvre2011substrate,brousseau2011substrate,seguin2020transparency,mufwene2002socio,kouwenberg2008problem, farquharson2012african}. 
Because we restrict our focus based on these linguistic traits, we also include some African Creole languages that are close phylogenetic relatives to our American focus languages, with some linguists arguing for a common ancestor \cite{mcwhorter2000missing}.\footnote{We also include three distinct control languages: Sango and Kituba (African Creole languages with morphosyntactic proximity to Niger-Congo languages but no lexical proximity to European languages), and Tok Pisin (an Oceanic Creole language with lexical proximity to a Germanic language---English---but no relation to Niger-Congo languages.)} 
% with the same characteristics.  According to prevalent linguistic theories,
For instance, Sierra Leonne's Krio is likely related to Maroon Creole languages like Ndyuka and Saramaccan \citep{bhatt2012structure}. 
Linguists have long noted the linguistic proximity of Louisiana Creole and French Guianese Creole to Creole languages of the Indian ocean (Mauritian, Seychellois, and R\'{e}union Creole) \cite{papen1978french}. 
Papiamento is considered a descendant of Kabuverdianu by some \cite{romero2010language}, and Jamaican has the same phylogenetic relatives as Ghanaian Pidgin \cite{amoako1992ghanaian,cassidy1966multiple}.

% \subsubsection{Additional language info - remove this}

% \nrr{Not sure if we actually want most of what's in this subsubsection. I originally envisioned it, but I think it will bore NLP folks.}

% Many Creole languages developed in the Americas as a result of colonialism.

% \nrr{individual language info: these should be ordered by number of native speakers}

% \paragraph{Haitian}
% is the first official language of the Republic of Haiti. It has over 11 million native speakers worldwide \cite{bartens2021making}, including a diaspora of over 1 million living outside Haiti \cite{audebert2017recent}. Haitian has been a language of global interest in recent years, due in part to immigration, refugee populations, and natural disasters \cite{heinzelman2010crowdsourcing, margesson2010haiti, rasmussen2015development}. Haitian is morphosyntactically related to West African languages in the Gbe and Kwa subfamilies \cite{lefebvre2011substrate,brousseau2011substrate,seguin2020transparency,kouwenberg2008problem}. 

% \paragraph{Jamaican Patois} or Jamaican is the lingua franca of Jamaica. It has 3.2 million speakers, according to Ethnologue. Jamaican is morphosyntactically related to West African languages in the Gbe, Kwa, Igboid, Yoruboid, and Bantu subfamilies \citep{mufwene2002socio, kouwenberg2008problem, farquharson2012african}.

\subsection{Collection methods}
\label{sxn:col_meth}

\begin{table*}
    \centering
    \small
    \begin{tabular}{
        rrrrr
        rrr
    }
        \toprule
        \multicolumn{5}{c}{\textbf{Bitext}} & \multicolumn{3}{c}{\textbf{Monolingual}}\\
        \cmidrule(lr){1-5}
        \cmidrule(lr){6-8}
        & 
        \multicolumn{2}{c}{\textit{Web}} & \multicolumn{2}{c}{\textit{PDF}} & & & \\
        \cmidrule(lr){2-3}
        \cmidrule(lr){4-5}
        \textit{Prev. pub.} & \textit{aligned} & \textit{articles} & \textit{aligned} & \textit{other} & \textit{Prev. pub.} & \textit{Web} & \textit{PDF} \\
        \midrule
        14196475            & 21963            & 216756            & 18614            & 4683           & 2767602             & 607657       & 27081   \\ 
        \bottomrule
    \end{tabular}
    \caption{Number of segments gathered from each source type/extraction method.}
    \label{tab:data_collec}
\end{table*}

We divide our dataset collection methodology into two stages: \textit{gathering} and \textit{extraction}. 
\num{25} Creole languages were selected for an active \textit{gathering} effort (\uline{underlined} in Table~\ref{tab:langs}). 
We excluded Haitian because its data were abundant in already identified sources. 
Further data were found for \num{17} other languages, also included in Table~\ref{tab:langs}. 
For each of the \num{25} \textit{gathering} languages, we performed the following steps. 
\textbf{First}, we searched research databases using query templates to track down already curated datasets. 
We searched “\texttt{[language name]}” on the ACL Anthology,\footnote{\url{https://aclanthology.org}} followed by “\texttt{[language name]} machine translation”, “\texttt{[language name]} NLP”, and “\texttt{[language name]} translation” on Google Scholar.\footnote{\url{https://scholar.google.com}} 
Query results from these search engines were typically prohibitively many, so we browsed top results until it became clear that the remainder were no longer relevant. 
(Individual data gatherers used best judgment for each language.) 
For some languages we multiplied queries to accommodate for alternate language names. 
(See Table~\ref{tab:lang_aliases} in Appendix ~\ref{sxn:app1}.)
\textbf{Second}, for each language, we checked each of the following databases for parallel or monolingual corpora: OPUS \citep{tiedemann-2012-parallel}, Oscar \citep{OrtizSuarezSagotRomary2019} and LDC.\footnote{\url{https://www.ldc.upenn.edu}}
% and OpenSubtitles\footnote{\url{https://opus.nlpl.eu/OpenSubtitles2018.php}} \citep{lison-tiedemann-2016-opensubtitles2016}.
\textbf{Third}, we scoured the web for books with translations or additional resources. 
\textbf{Fourth}, we contacted researchers in the languages' speaking communities for leads to potential data sources. 
Though we attempted various methods of contact whenever possible, response rates were low for this step. 
In all we gathered 107 sources such as anthology websites promoting cultural heritage or language revitalization, educational materials, and government documents. 

After completing \textit{gathering}, we moved to \textit{extraction}. 
At this stage, we divided each of the gathered resources into groups, based on the data format. 
The six groups were: (1) parallel data previously published as a bitext, (2) web sources with aligned parallel sentences, (3) web sources containing unsegmented articles of text with translations, (4) PDF sources with aligned parallel sentences, (5) other PDF sources, and (6) sources of monolingual data. 
The amount of segments for each of these groups is summarized in Table~\ref{tab:data_collec}, with a breakdown per language in Table~\ref{tab:app_dtype} (Appendix~\ref{sxn:app1}).

We immediately consolidated the previously published bitexts, including single-language resources from LAFAND-MT \citep{adelani-etal-2022-thousand}, KreolMorisienMT \citep{dabre2022kreolmorisienmt} and the Caribe Trinidadian-English dataset \citep{smith2022trinidad}. 
We gathered monolingual data from the MADLAD-400 clean corpus \citep{kudugunta2023madlad} and JamPatoisNLI \citep{armstrong2022jampatoisnli}. 
To create novel bitexts, we then used the BeautifulSoup\footnote{\url{https://www.crummy.com/software/BeautifulSoup/}} and Selenium\footnote{\url{https://www.selenium.dev}} Python packages to extract text from web sources and the PyPDF2\footnote{\url{https://https://github.com/py-pdf/pypdf}} package for PDF sources. 
Organizing bitexts from sources with aligned parallel sentences was generally straightforward. 
When only unsegmented articles with translations were available, we segmented and aligned text based on punctuation and then manually corrected errors. 
Manual correction was performed by data extractors with sufficient proficiency in the languages involved. 
Our diverse team has some proficient speakers in our target languages (one L1 Martinican Creole speaker, two L2 Haitian speakers, one L2 Guadeloupean Creole speaker, one L2 Louisiana Creole speaker, two L2 Naija speakers, and one L2 Ghanaian Pidgin speaker). 
Though this did not cover anywhere near all of our included languages, proximity between Creole languages and related languages made the task doable.\footnote{For instance, one Louisiana Creole speaker and one Haitian speaker---both also proficient in English, French, and linguistics---were able to manually correct alignment for Seychellois-English bitexts, due to Seychellois’ lexical overlap with Haitian, Louisiana Creole, and French.}
In general, the data extraction details varied for each source, since PDFs and websites have a wide variety of individualized styles and formats. 
In all, we extracted data from 39 of the 107 sources from our \textit{gathering} stage.\footnote{The remainder contained data in less accessible formats that we deemed too tedious to include in this work. We intend our dataset to be a living resource and hope to continually add to it as we encounter and format more data for Creole languages.} 
Attributions for all data sources are in Appendix~\ref{sec:attribution}. 

As an appendage to our methodology, we incorporate data from published multilingual resources, including all bitexts from CreoleVal \citep{lent2023creoleval}, Lego-MT \citep{yuan-etal-2023-lego}, FLORES-200 dev and NLLB train data \citep{team2022language}, and AfricaNLP'23 \citep{robinson2023african} for any of our focus languages.\footnote{We ensure not to include any data labeled for testing in our own data for training or tuning.}  
We retrieved Bible translations from JHU \citep{mccarthy-etal-2020-johns} for 18 languages, with up to four unique translations for each. 
We selected the four fullest English and French Bibles with reasonably modern text style from the same corpus, and formed up to eight bitexts for each language. 
We also contribute 360K previously unreleased unique parallel sentences from the Church of Jesus Christ of Latter-day Saints for \texttt{hat}, \texttt{pap}, and \texttt{tpi} with English. % Can we add full names? Do we have space?
This data comes from religious sources, including scripture, instruction, discourses, humanitarian resources, genealogy, and administrative documents. 
We scraped small parallel corpora in the educational domain from APICS \citep{apics} for \num{39} languages. 
We aggregated further published bitexts by crawling the web via Python's \texttt{mtdata} package, which points to data from multiple online sources. 
Last, we added monolingual Wikipedia dumps for \texttt{jam}, \texttt{gcr}, \texttt{gpe}, and \texttt{srn}.

\subsection{Data amounts}

\begin{table*}
    \centering
    \small
    \begin{tabular}{
        rccrccrcc
    }
        \hline
        \multicolumn{3}{c}{ } & \multicolumn{3}{c}{ } & \multicolumn{3}{c}{ }\\
        & \textbf{Max.} & \textbf{Ours} & & \textbf{Max.} & \textbf{Ours} & & \textbf{Max.} & \textbf{Ours} \\
        & \textbf{prev.} & (pub. / all)  & & \textbf{prev.} & (pub. / all)  & & \textbf{prev.} & (pub. / all) \\
        \cmidrule(lr){1-3}
        \cmidrule(lr){4-6}
        \cmidrule(lr){6-9}
        \textbf{acf} & 15989 & 4406 / 23916 & \textbf{gcf} & 96 & 6467 / 6467 & \textbf{mue} & {-} & 147 / 147  \\ 
        \textbf{aoa} & {-} & 198 / 198 & \textbf{gcr} & {-} & 1433 / 1433 & \textbf{pap} & 4898029 & 4968965 / 5363394  \\ 
        \textbf{bah} & {-} & 327 / 327 & \textbf{gpe} & {-} & 223 / 223 & \textbf{pcm} & 31128 & 8084 / 47455  \\ 
        \textbf{brc} & {-} & 222 / 222 & \textbf{gul} & 7990 & 266 / 8831 & \textbf{pov} & {-} & 480 / 480  \\ 
        \textbf{bzj} & 23406 & 229 / 31002 & \textbf{gyn} & {-} & 258 / 258 & \textbf{pre} & {-} & 243 / 243  \\ 
        \textbf{bzk} & {-} & 391 / 391 & \textbf{hat} & 4256455 & 5715227 / 6023034 & \textbf{rcf} & {-} & 285 / 285  \\ 
        \textbf{cab} & 20879 & {-} / 20879 & \textbf{icr} & 15702 & 317 / 16774 & \textbf{sag} & 262334 & 260560 / 535310  \\ 
        \textbf{cri} & {-} & 306 / 306 & \textbf{jam} & 25206 & 434 / 28713 & \textbf{srm} & 42303 & 440 / 59053  \\ 
        \textbf{crs} & 222613 & 3186 / 225875 & \textbf{kea} & 129449 & 132931 / 132931 & \textbf{srn} & 583830 & 6620 / 615010  \\ 
        \textbf{dcr} & {-} & 189 / 189 & \textbf{kri} & 50438 & 185 / 66736 & \textbf{svc} & {-} & 321 / 321  \\ 
        \textbf{djk} & 45361 & 15266 / 68833 & \textbf{ktu} & 7886 & 175 / 10737 & \textbf{tpi} & 424626 & 451758 / 925648  \\ 
        \textbf{fab} & {-} & 204 / 204 & \textbf{lou} & {-} & 1860 / 1860 & \textbf{trf} & {-} & 1691 / 1691  \\ 
        \textbf{fng} & {-} & 160 / 160 & \textbf{mart1259} & {-} & 5153 / 5153 & \textbf{wes} & {-} & 223 / 223  \\ 
        \textbf{fpe} & {-} & 259 / 259 & \textbf{mfe} & 191909 & 25633 / 233320 &  &  &   \\
        \hline
    \end{tabular}
    \caption{Size of largest bitext data collected for Creole languages to date, compared with our full bitext sets (all) and their public subsets (pub.).
    Bitext size is measured as the number of unique Creole language sentences paired with a translation in any target language.
    }
    \label{tab:data_amt2}
\end{table*}

Table~\ref{tab:data_amt2} compares our own data size with the largest previously collected dataset for each language. Ours is the first dataset for \num{21} languages, and the first public dataset for \num{9} more. (See Table~\ref{tab:data_amt} for a comparison with individual prior works.)

Our datasets are, to our knowledge, the largest ever collected for each of these Creole languages, (1) because ours contain a conglomerate of the compared previously disparate sets, and (2) because of the additional data previously inaccessible to MT that we gathered in our \textit{extraction} process. 
These latter novel data make up \qty{262}{\kilo\quantity} of the bitext sentences in our dataset.
% \S\,\ref{sxn:col_meth} goes over our data extraction methods. 
Figure \ref{fig:maps} illustrates the size of each language's dataset, with circles at the coordinates for each language from Glottolog \citep{nordhoff2011glottolog} (zoomed in on Latin America). 
Filled blue markers have area corresponding to the square root of the bitext size. Hollow red markers indicate languages for which we found no bitext data (\texttt{scf}, \texttt{kmv}).

\subsection{Diversity of genres}
\label{sxn:genres}

\begin{figure}[t]
    \centering
    \includegraphics[width=1\linewidth]{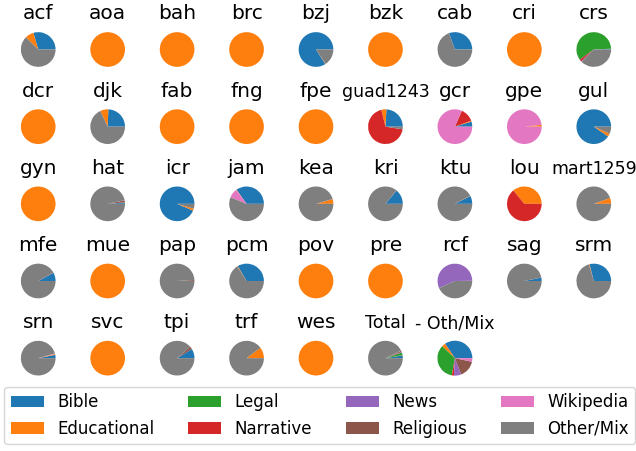}
    \caption{Genre proportions of each language's data (bitext and monolingual). We exclude the "Other/Mix" genre in the final pie to filter out large NLLB sets and show no majority among other genres.}
    \label{fig:data_genre}
\end{figure}

As mentioned in \S\,\ref{sxn:rw}, prior works in Creole language MT are severely restricted in terms of genre, prohibiting them from general-purpose MT.
We mitigate this by including a diversity of genres. 
Figure~\ref{fig:data_genre} shows that more than half of languages (26/41) cover at least two genres.\footnote{Table~\ref{tab:app_genre} in Appendix~\ref{sxn:app1} contains the same information in colorblind-friendly numerical form.} 
The "Other/Mix" genre dominates charts for Haitian, Papiamento, and Total, because of the large NLLB train sets (sets we do not even include in our model training, see \S\,\ref{sxn:exp}). 
We include a final chart indicating the total with "Other/Mix" excluded, showing no majority among other genres. 
We acknowledge that our dataset's genre diversity can still be improved in future iterations, but we highlight the dramatic increase in diversity it offers, compared to previous work. 

\section{Modeling Experiments}
\label{sxn:exp}

We now describe the experimental setup to show the utility of our datasets by training MT systems. 
We train and release models on our full datasets, but we also train on only the data we publicly release, to show its stand-alone utility. 
We experiment with cleaning our train sets versus leaving them as-are, and in some experiments we fine-tune mBART \citep{liu-etal-2020-multilingual-denoising} rather than training from scratch. 

\subsection{Dataset Preprocessing}
\label{sec:preprocessing}

\noindent \textbf{Two collections}: We maintain two primary data collections: one with all available data, henceforth called \textbf{all}; and a subset with all publicly releasable data, henceforth called \textbf{public}. 
To ensure the comparability of our models with previous work, we do not modify already available test splits. 
Moreover, for our experiments, we ensure that there is no overlap between these pre-existing test sets and our own data, by removing from our train/dev splits every sentence pair where either the source or the target is present in a pre-existing test set.\footnote{This is made necessary by the variety of sources we used but results in a loss of less than \qty{1}{\percent} of the overall sentence pairs and allows for comparisons with the state-of-the-art that are as fair as possible.}

\noindent \textbf{Evaluation splits:} After filtering, we prepare a train/dev/test split for each language pair of the remaining data by aggregating all sentences and splitting randomly with a fixed random seed, and a target ratio of \qty{85}{\percent}\,/\,\qty{5}{\percent}\,/\,\qty{10}{\percent}, with minimum \num{50} and maximum \num{2000} sentences for the dev and test sets. 
We discard train sets with fewer than \num{100} sentences. 
We still conduct zero-shot evaluations for language pairs for which training data was removed (indicated with an `*' in Figure~\ref{fig:res}). 
We exclude NLLB training sets for \texttt{hat}, \texttt{kea}, \texttt{pap}, \texttt{sag}, and \texttt{tpi} due to their overbearing size and observed poor quality, reducing our train set by 10.6M parallel sentences to a size of 450K.  
%Finally, we discard all training pairs that do not include either English or French, most of them being so small that it doesn't affect the data amount much, but simplifies the modeling. 
% Note that we discard data only pertaining to training and not for testing, which implies that we have zero shot evaluation for some language pairs.

\noindent \textbf{Cleaning:} Each dev and test set was cleaned according to the \href{https://github.com/GILT-Forum/TM-Mgmt-Best-Practices}{GILT Leaders Forum’s Best Practices in Translation Memory Management}.\footnote{We release cleaning software on our repository \github{}.} 
We removed segments that were: empty, containing unbalanced brackets, mostly non-alphabetic characters, containing a source the same as the target, fewer than \num{3} words, and containing a higher number of characters than \num{5} standard deviations above the mean for data of that language. 
We normalized escaped characters and entities, white spaces, quotation marks, and the overall character set for each language. 
We removed any spurious characters that do not contribute semantically or syntactically in a segment and remove duplicates for all segments after cleaning to ensure there is no development and test set contamination. 
We toggle whether we thus clean train sets in both \textbf{public} and \textbf{all}, leading to a total of four configurations: \textbf{public}, \textbf{public-cleaned}, \textbf{all}, \textbf{all-cleaned}. 
Note that the dev and test sets, and hence models and their results, between \textbf{public} and \textbf{all} are not comparable, but those among cleaned and non-cleaned versions are (since the former variable implies independently split test sets, but latter variable toggles only whether \textit{train} data were cleaned).

\subsection{Implementation and Training}

We train models with YANMTT \cite{dabre-etal-2023-yanmtt}, a toolkit that supports multilingual training and fine-tuning of mBART-50 \cite{tang-etal-2021-multilingual} (minute model details in Appendix~\ref{sec:train-details}).
We first train from \textbf{scratch}, using all train data to fit a multilingual SentencePiece tokenizer \citep{kudo-richardson-2018-sentencepiece} of \qty{64}{\kilo\quantity} subwords for all languages. 
Given our four experimental configurations, we train four such models. 
Next, we fine-tune many-to-many \textbf{mBART}-50 \citep{tang-etal-2021-multilingual}, repurposing the language indicator tokens for our Creole languages for simplicity. 
To keep a manageable computation budget for the compute-hungry mBART models, we only train them on the \textbf{clean} data configurations. 
We decode translations with beam size \num{4} and length penalty \num{0.4}.

% \nrr{Basics here @Raj - YANMTT and hyperparameters, tokenization details}

% \nrr{Then explain the methodological variations @Raj -- full and then public cleaned / uncleaned, mBART for hat--eng}

\section{Results and Discussion}
\label{sec:bench}
% We now analyze the results of our experiments.

\begin{figure*}[t]
    \centering
    \includegraphics[width=\textwidth]{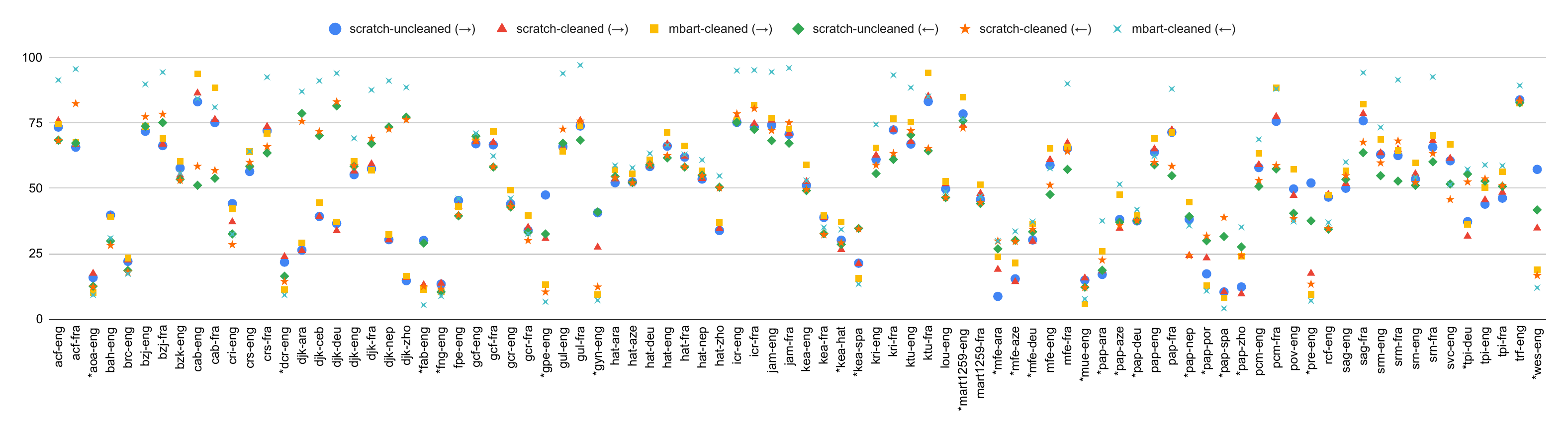}
    \includegraphics[width=\textwidth]{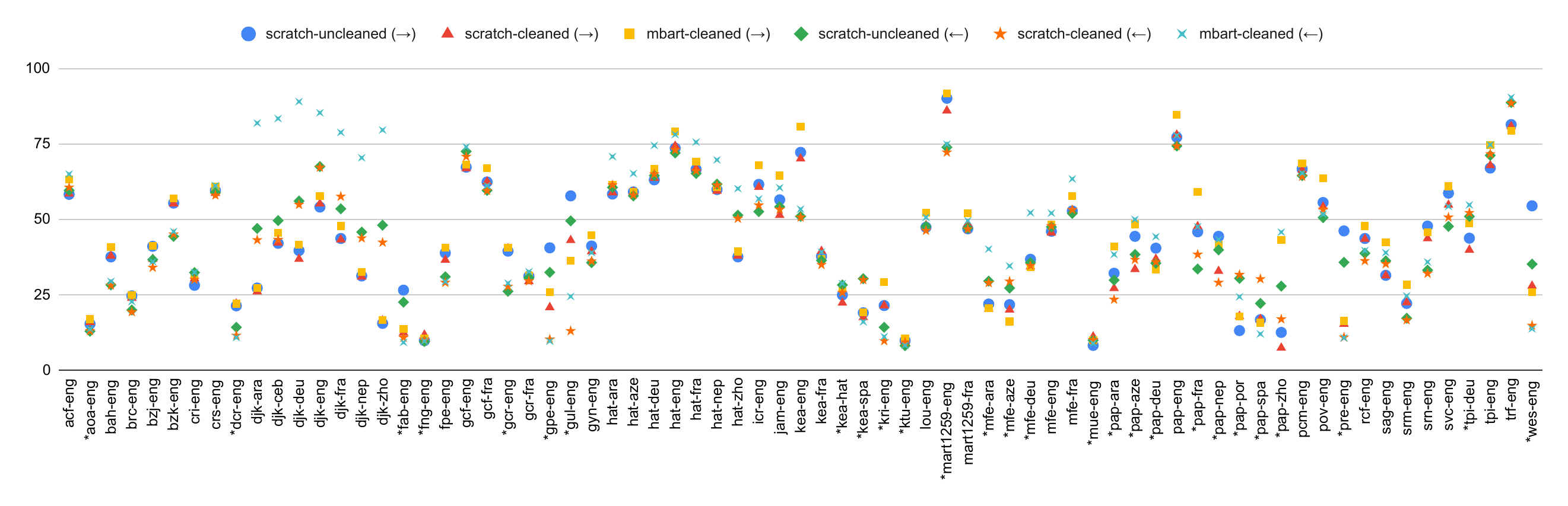}
    \caption{chrF scores on our newly created test sets using models trained on the \textbf{all} (top) and \textbf{public} (bottom) splits of our datasets. Given X-Y pair, $\rightarrow$ and  $\leftarrow$ represent the X to Y and Y to X translation, respectively. Zero-shot pairs are marked with an `*' sign.}
    \label{fig:res}
\end{figure*}

% \subsection{Main Results}
Figure~\ref{fig:res} shows the performance of all models, tested for all available language pairs. 
We use chrF \cite{popovic-2015-chrf} because previous studies \cite{mathur-etal-2020-tangled, rei-etal-2020-comet} show that it correlates better with human judgments than BLEU \cite{papineni-etal-2002-bleu}, and because semantic metrics such as COMET \cite{rei-etal-2020-comet} and BLEURT \cite{sellam-etal-2020-bleurt} lack support for low-resource languages.\footnote{Other recent low-resource MT works also prioritize chrF \cite{robinson-etal-2023-chatgpt, dabre-sukhoo-2022-kreolmorisienmt}.}
Because many MT researchers have a more intuitive grasp of BLEU than chrF, we also include BLEU scores in Figure~\ref{fig:res-bleu} of Appendix~\ref{sec:extra-results}. 

As expected, high-resource pairs exhibited better translation quality than their lower resource counterparts, with some deviations from this trend possibly due to inevitable non-uniformity in test set genre. 
Some possibly overfit pairs like Gullah-French scored chrF up to 97.2, and some zero-shot pairs like Papiamento-Spanish scored as low as 4.1. 

Despite the generally lower performance on their languages, we highlight the potential value of our dataset's smallest bitexts. 
As stated in \S\,\ref{sxn:intro}, Creole languages’ linguistic relationships open the possibility of powerful cross-lingual transfer. 
Previous studies \cite{Ernštreits_Fišel_Rikters_Tomingas_Tuisk_2022,10.1145/3406095,arivazhagan2019massively,team2022language} have shown that MT models trained on numerous languages can often be adapted to translate new languages with few examples. 
Our results corroborate this finding; despite having only 391 parallel sentences to split for Miskito Coast Creole (\texttt{bzk}), our models achieved BLEU and chrF up to 43.7 and 60.4, respectively, translating into English.
Vincentian Creole (\texttt{svc}) performed even better: up to 55.5 BLEU and 66.9 chrF into English, despite only 321 parallel sentences total. 
(Such scores would likely not be attainable via traditional bilingual training on such small sets \cite{arivazhagan2019massively}.) 
Strikingly, we perceived even higher performance on zero-shot Martinican-to-English translation, with 69.9 BLEU and 84.9 chrF. 
We hope future studies will reveal more about our dataset’s potential for low-resource cross-lingual transfer, and how it interfaces with dataset genre and diversity. 
This is an important exploration, not only to engineer better low-resource language systems, but as a scientific inquiry with general implications for low-resource language technologies. 

\noindent \textbf{Impact of corpora cleaning:} 
In Figure~\ref{fig:res}, regardless of the use of \textbf{public} or \textbf{all} data, models trained on cleaned data (triangles and stars) typically outperform their counterparts trained on non-cleaned data (circles and diamonds), for higher resource languages. 
Cleaning eliminates noise and reduces variability in the especially noisy Creole datasets, which helps translation quality. 
However, cleaning can hurt performance by reducing the already scarce data for the lower-resource, more noisy corpora (like Pichi-English with 259 sentence pairs). 

\noindent \textbf{Does fine-tuning help?} 
From Figure~\ref{fig:res}, a comparison of 
\textbf{scratch} and \textbf{mBART} models trained on cleaned data (squares and x's) reveals that mBART fine-tuning helps both \textbf{public} and \textbf{all} data training. 
This fine-tuning helps more for out-of-Creole directions, with an average improvement of 8.1 chrF over \textbf{scratch} models on \textbf{all} data (compared to 1.4 chrF for into-Creole). 
\textbf{Scratch} models converged after 500k training steps of 33k-token batches, and \textbf{mBART} fine-tuning needed 500k-600k steps of only 8.2k-token batches. 
While this implies that \textbf{mBART} is data efficient, \textbf{mBART} fine-tuning is slow due to training about 700\% more parameters. 
In total, \textbf{mBART} fine-tuning needed up to a week whereas \textbf{scratch} training needed only up to two days on the same hardware.

\begin{figure}[t]
    \centering
    \includegraphics[width=\columnwidth]{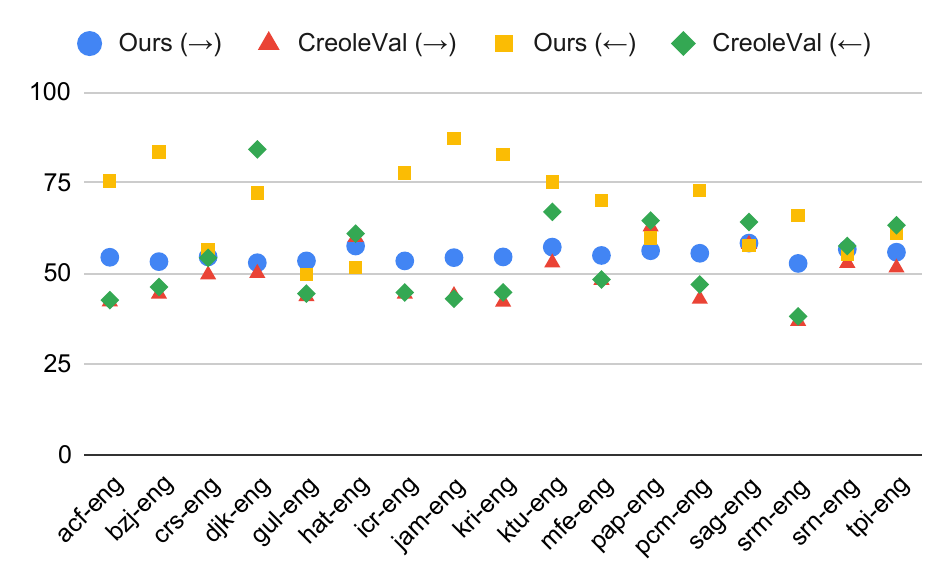}
    \caption{chrF for our model by fine-tuning \textbf{mBART} on \textbf{all cleaned} compared with CreoleVal. Given X-Y pair, $\rightarrow$ and  $\leftarrow$ represent the X to Y and Y to X translation, respectively.}
    \label{fig:ours-vs-creoleval}
\end{figure}

\noindent \textbf{Comparison on existing test sets:} 
Figure~\ref{fig:ours-vs-creoleval} compares our best models against results reported by CreoleVal \citep{lent2023creoleval} on relevant language pairs from the CreoleVal test sets. Since CreoleVal data is not public, we consider it apt to evaluate our best model (\textbf{mBART}) fine-tuned on \textbf{all cleaned} data. 
Overall, our models surpass CreoleVal models with +6.4 average chrF X$\rightarrow$ENG and +14.1 average chrF ENG$\rightarrow$X. 
Our dataset is much larger and more domain diverse than CreoleVal's; hence our improved results on most language pairs show that increasing data tends to be beneficial even if it reduces the domain specificity. 
We note that on a few language directions (8 of 34), CreoleVal's model still beat ours, indicating the possibility for negative interference from a more diverse train set in some instances. 
In Appendix~\ref{sec:extra-results} we provide some additional comparisons of bilingual versus multilingual fine-tuning and evaluations on other existing benchmarks: Lego-MT \cite{yuan2023lego} and KreolMorisienMT \cite{dabre2022kreolmorisienmt}. 

% \nrr{Cite chrF \cite{popovic-2015-chrf} and ref Figure~\ref{fig:res}}

% \nrr{@NATE: add figures soon! Multiple tables in this section - or one bar chart???}

% \nrr{
% Outline:

% \begin{itemize}
%     \item Cleaned data is always better than noisy data for our in-domain test sets
%     \item However, matching the domain of the test set yields best performance. (Noisy data did best on Lego-MT test sets) 
%     \item mBART fine-tuning helps performance for higher resource language pairs (mention that mBART was trained on fewer iterations)
%     \item Our model performs better than CreoleVal 4.9 chrF average on X$\rightarrow$ENG directions, and performs on par (-0.2 chrF average) for ENG$\rightarrow$X.
% \end{itemize}
% }

% \section{Impact on/of Communities}

% \nrr{Think we should combine this with the Conclusion section!}

% \nrr{Some points about "othering" Creole langs}

% \nrr{Acknowledge some groups may not want MT -- for Ethics section}

% \nrr{Anything we can say about communities}

% \nrr{Add disclaimers to Ethis sxn: might have sensitive info, some sentences might be offensive in certain cultural contexts, add disclaimers about engaging communities, one person not representing community}

\section{Conclusion}

In this work, we compile the most comprehensive dataset to date for MT of Creole languages in the Americas. 
By aggregating disparate previous works and incorporating new data sources via scraping the web and PDFs, we expand MT datasets to 21 new languages and produce the largest and most genre-diverse in 20 more. 
We release translation models in 172 language directions, with 26/32 beating state-of-the-art benchmark performance, as well as a public dataset with 11.6M aligned bitexts and 3.4M monolingual sentences. 

A large multilingual bitext like ours has potential to build the best yet or first ever MT models for many languages, something we accomplished on a surface level in this work but hope future works will continue. 
The data present a number of other potential uses, including: (1) training language models for applications like spelling correction \cite{abdulrahman-hassani-2022-language, etoori-etal-2018-automatic, aljefri2013contextieee}; (2) availing textual data for applications like speech recognition and speech translation, which can be vital to low-literacy communities \cite{robinson2022tts, gao2021pre, rossenbach2020generating}; (3) potential to study cross-lingual transfer between Creole languages and their phylogenetic relatives in yet unseen depth \cite{robinson2023african}; (4) research on the effects of linguistic data augmentation for small MT datasets (focusing on Creole languages' unique linguistic position); (5) development of MT-assisted documentation tools for those languages that are endangered \cite{bird-chiang-2012-machine}; (6) the introduction of a common repository where any researchers in Creole NLP can accumulate datasets together and advance in collaboration \cite{lent2023creoleval}; etc. 
We hope that this work will provide valuable translation technologies to communities that have been historically under-served and inspire community-oriented efforts to further expand work on these low-resource languages.

\section*{Acknowledgements}

We thank Suzanna Sia, Chris Emezue, Heather Lent, David Mortensen, Oliver Mayeux, Michael Gisclair, Arya McCarthy, Ruth-Ann Armstrong, Carter Charles, Jeff Allen, Jamell Dacon, and Fritz-Carl Morlant for their contributions to the ideas and processes of this work.

\section*{Limitations}

%\nrr{Add something about unorthodox orthographies! and speech technologies}
Across languages, writing systems change over time due to linguistic changes, such as the loss of distinctions between sounds; or metalinguistic changes, such as the desire to associate a speech community with a more prestigious one (or conversely show that the speech community is distinct). 
This concern is exacerbated for Creole languages, which tend to have very recent and often still developing standardization processes \citep{deuber2007dynamics, valdman2005vers, rajah2009use}.
For several of our languages, especially Louisiana Creole and French Guianese Creole, we rely extensively on texts that were written more than 100 years ago and thus use spelling systems that have been partially or wholly superseded by new systems. 
In principle, it is possible to rewrite such texts with more recent conventions using language-specific scripts, but we opted to use the original orthographies for this work to keep the processing pipelines as similar as possible across languages. 
%\nrr{Can't have all the data - living dataset, expand upon}
In the course of our data collection efforts, we identified several sources which we did not have time to process. In the future, we intend to continue adding new sources into our dataset, with a preference for those which are already publicly available.
%\nrr{Compute restricted}

% \nrr{Need for human annotations}

% \nrr{Need to train mBART longer}

Currently in our experiments, the newly created development and test scores are not comparable across the \textbf{public} and \textbf{all} splits due to their independent splitting processes (and hence dissimilar test sets). 
Our future work will focus on having development and test sets that are common across both splits, regardless of whether the training data in said splits are cleaned or not, for consistent comparisons.

%\nrr{Acknowledge this is an ongoing work of data collection. -- put this in Limitations}

\section*{Ethics Statement}

Because Creole languages have frequently been the target of "othering" and marginalization \citep{degraff2005linguists,lent2023creoleval}, it is important to approach Creole language technologies with sensitivity. From a linguistic standpoint, the question of "What exactly makes a language Creole?" has contribued to the lack of prestige that many Creole languages currently face. For this study, we use existing literature to identify languages that have been considered Creole, but do not seek to assert a singular "Creole essence."

As with any other linguistic community, there are considerable differences in opinion concerning the desirability of MT in various Creole-speaking communities. We acknowledge that MT technologies do not inherently benefit all Creole language speakers. Many of them can already use existing MT tools in a different language, lessening the immediate benefit of MT tools in the relevant Creole. Others may be concerned about machine translation displacing human translators, or view the entire concept of MT as offensive, as it directly broaches the subject of linguistic differences between their languages and European languages (differences which are still broadly stigmatized). We also acknowledge that the intended use of some of the resources we collected may not have been MT. We do not wish to undermine the original purposes of anthology-style data but hope our work will support these endeavors. 

We also acknowledge that the texts we have assembled, especially those which are older, religious and/or political, reflect many different viewpoints that may be considered dated, contested, or offensive in some cultural contexts. This is a natural part of the data collection process, but it is not an endorsement of the content of any given text. We did not seek to remove such viewpoints from our data, as they are culture-specific. Another risk of data collection is the inclusion of personally identifiable information that may pose a risk to some users. This is a particular problem with a commonly used Haitian-English bitext from WMT 2011 \citep{callison2011findings}. Though it is difficult to avoid data contamination in this vein completely, we avoid including this dataset to mitigate this risk. 
We also acknowledge the potential for bias in our dataset, since it is not perfectly balanced in terms of genres and topics. 
We encourage more application-oriented work in the future to report MT results broken down by test set genre. 

In our conception, the primary beneficiaries of Creole MT technologies would be monolingual Creole language speakers. Many monolingual Creole language speakers have limited literacy and would perhaps benefit from speech translation systems more than text-based systems. As such, we encourage future work in the area of speech technologies for Creole languages and hope the textual materials and models we provide in this work can be of use to that end.

Lastly, ChatGPT was used to assist software writing for this project. We acknowledge the ethical implications of LLM use are still being understood.

\bibliography{anthology,custom}

\clearpage

\appendix

\section{Additional Data Information}
\label{sxn:app1}

% Table~\ref{tab:app_genre} contains numerical values corresponding to Figure~\ref{fig:data_genre}. 
% Table~\ref{tab:app_dtype} shows what formats we retrieved data from for each language (summarized in Table~\ref{tab:data_collec}). 
% Table~\ref{tab:data_amt} shows our dataset amounts compared with previously collected datasets. Table~\ref{tab:lang_aliases} shows alternate names for languages with more than one used in data gathering.

Table~\ref{tab:data_amt} contains bitext sizes for individual previous publications for each language, compared to our own datasets (summarized in Table~\ref{tab:data_amt2}). 
Table~\ref{tab:app_genre} provides numerical values for our dataset's genre composition (corresponding to Figure~\ref{fig:data_genre}). 
Table~\ref{tab:app_dtype} gives language-by-language details of the types of data we extracted in our collection methodology (summarized in Table~\ref{tab:data_collec}). 
Table~\ref{tab:lang_aliases} shows alternate names for languages with more than one used in data gathering.

\section{Experimental Details}
We describe some additional details related to training and evaluation. 

\subsection{mBART-50 Token Repurposing}
mBART-50 has 51 special tokens corresponding to exactly 51 languages in our experiments, of which 41 are Creoles. Therefore, we simply repurpose these tokens where we fix the tokens for English, French, Spanish and Portuguese to the corresponding tokens in the mBART-50 tokenizer and then randomly assigned other tokens to Creoles. We found that this only affects initial training, as the model has to re-learn that the token is used to translate into another language.

\subsection{Training and Convergence}
We train all models until convergence\footnote{We use annealing to declare convergence: we first wait until the model does not show BLEU improvements for \num{10} consecutive evaluations. We then decrease the learning rate by half and wait until the model does not show BLEU improvements for \num{15} evaluations. Finally, we decrease the learning rate by half again and if the model does not show improvements for \num{20} evaluations then we declare convergence.} on the validation sets, evaluated every \num{5000} steps, and up to a maximum of \num{500 000} steps. For the models trained from scratch, this took three days on a single node of eight \qty{32}{\gibi\byte} V100 GPUs. (Fine-tuning mBART took much longer, as noted in \S\,\ref{sec:bench}.) See Table~\ref{tab:hyperparams} for the sizes of our models.

% \nrr{Add model parameter sizes!}

\subsection{Hyperparameters}\label{sec:train-details}
\begin{table}[th]
    \centering
    \small
    \begin{tabular}{lr}
        % Hyperparameter & Value\\
        \toprule
        Optimizer & AdamW \\
        Learning Rate & \num{1e-3} (\num{3e-4})\\
        Weight Decay & \num{1e-5} \\
        \#encoder/decoder layers & \num{6} (\num{12})\\
        hidden size & \num{512} (\num{1024}) \\
        FFN size & \num{2048} (\num{4096}) \\
        \#parameters & \qty{77}{\mega\quantity} (\qty{611}{\mega\quantity}) \\
        data sampling temperature\tablefootnote{We selected 2.0 to oversample the smaller corpora only slightly. We noticed in our peliminary experiments that the standard temperature of 5.0 caused the higher resource pairs to suffer.} & \num{2.0}\\
        batch size\tablefootnote{This value here is the total batch size in tokens over eight \qty{32}{\gibi\byte} V100 GPUs.} & \num{32 768} (\num{8192}) \\
        dropout & \num{0.1} (\num{0.1}) \\
        label smoothing & \num{0.1} \\
        \bottomrule
    \end{tabular}
    \caption{Hyperparameter settings for models trained from scratch. Values in parentheses indicate those used for fine-tuning mBART-50.}
    \label{tab:hyperparams}
\end{table}

The relevant training hyperparameters are given in Table~\ref{tab:hyperparams}.

\subsection{Additional Models Trained}
We train a simpler version of \emph{m2m-mBART} focusing only on Haitian--English bidirectional translation to determine whether its better to focus on fewer language pairs during fine-tuning. The key changes in hyperparameters are learning rate (\num{3e-5}) and dropout (\num{0.3}).

\section{Additional Results}\label{sec:extra-results}

\subsection{BLEU Scores For Our Test Sets}

\begin{figure*}[t]
    \centering
    \includegraphics[width=\textwidth]{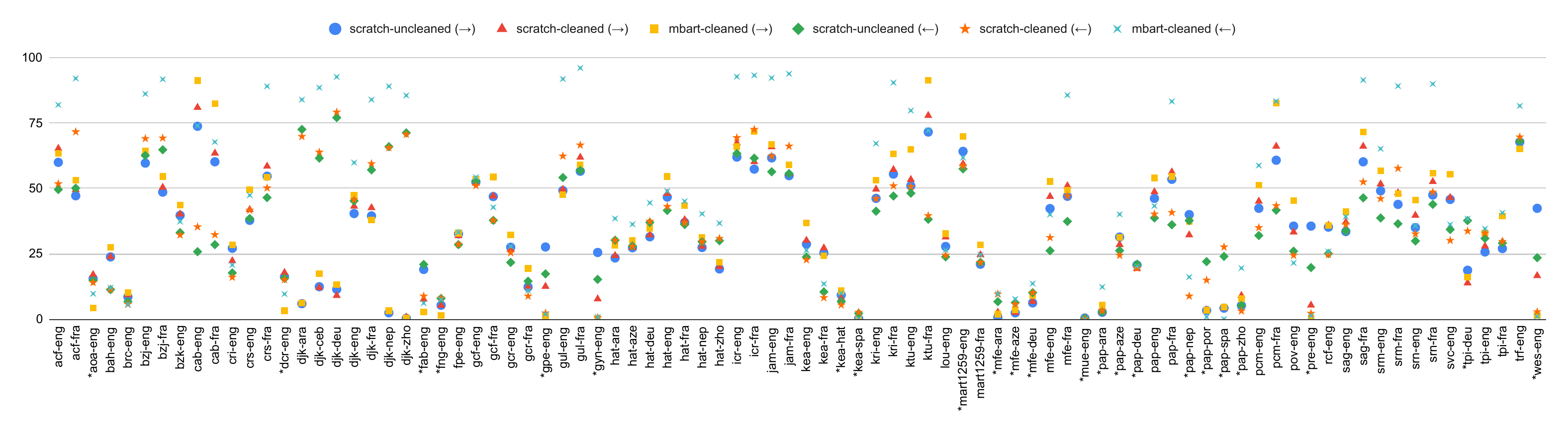}
    \includegraphics[width=\textwidth]{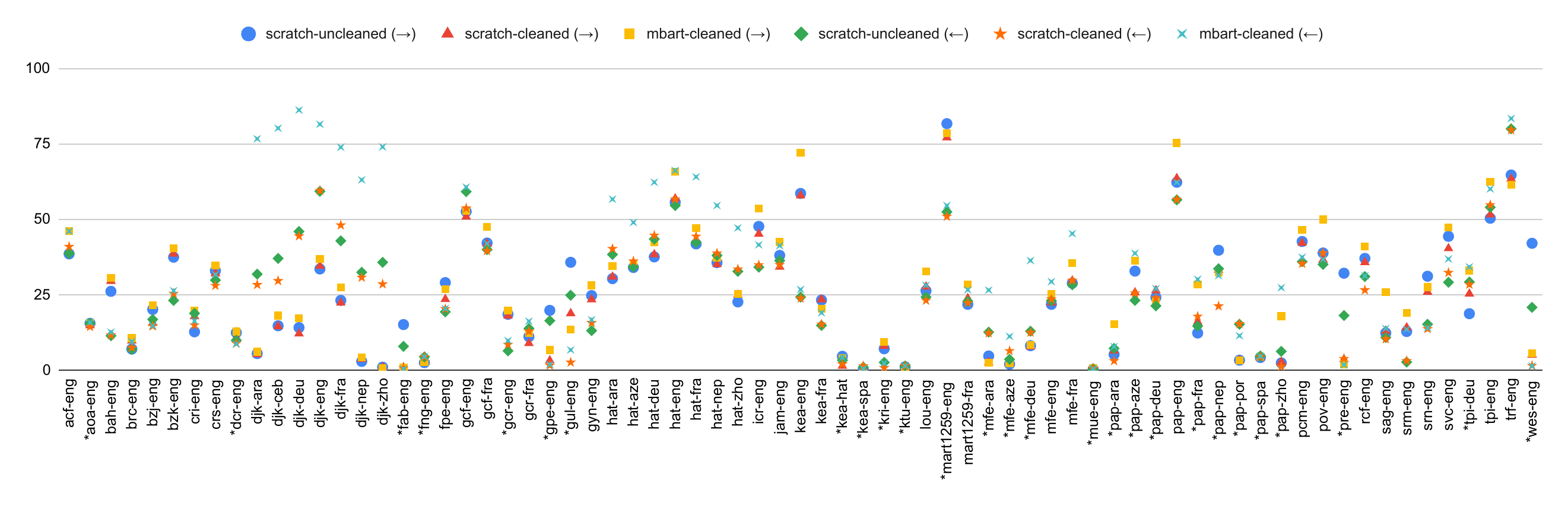}
    \caption{BLEU scores on our newly created test sets using models trained on the \textbf{all} (top) and \textbf{public} (bottom) splits of our datasets. Given X-Y pair, $\rightarrow$ and  $\leftarrow$ represent the X to Y and Y to X translation, respectively. Zero-shot pairs are marked with an `*' sign.}
    \label{fig:res-bleu}
\end{figure*}

Figure~\ref{fig:res-bleu} shows the BLEU scores on our newly proposed test sets. These are analogous to the chrF scores presented in Figure~\ref{fig:res}. 

\subsection{Bilingual vs Multilingual mBART fine-tuning}

\begin{figure}[th]
    \centering
    \includegraphics[width=\columnwidth]{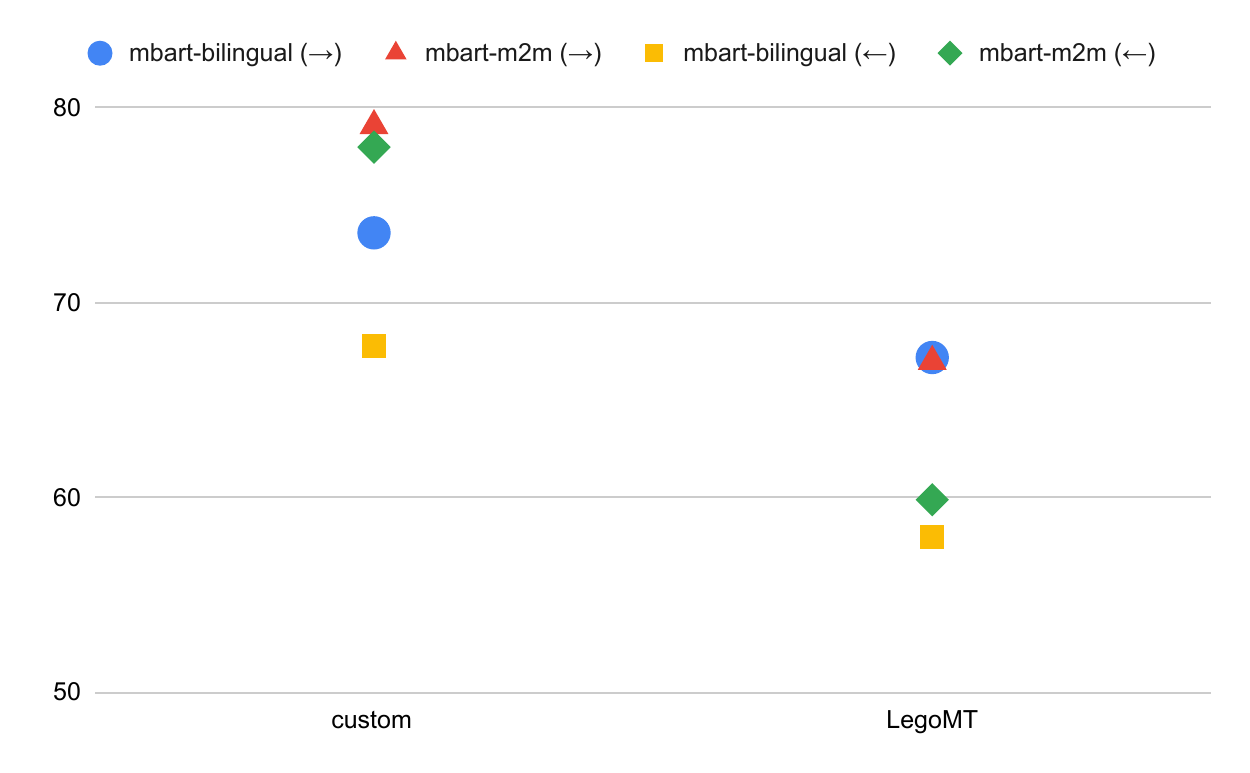}
    \caption{Comparing Haitian to English ($\rightarrow$) and  English to Haitian ($\leftarrow$) translation quality for bilingual and multilingual fine-tuning for our custom test set as well as the Lego-MT test set.}
    \label{fig:bi-vs-m2m}
\end{figure}

Figure \ref{fig:bi-vs-m2m} shows the results of fine-tuning mBART-50 only on Haitian--English vs on multilingual data, for the \textbf{public} split mentioned in Section~\ref{sec:preprocessing}. We can see that bilingual models are inferior to multilingual models. The same trend exists for the \textbf{all} split.

\begin{figure}[t]
    \centering
    \includegraphics[width=\columnwidth]{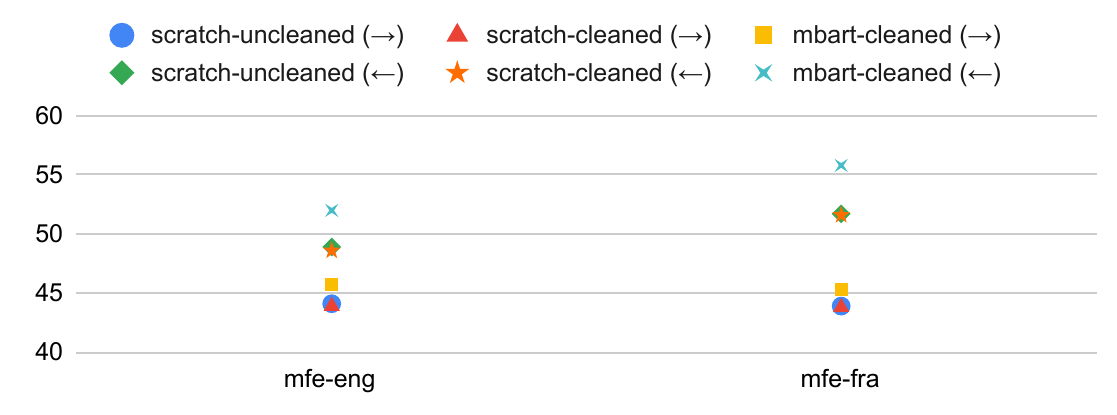}
    \includegraphics[width=\columnwidth]{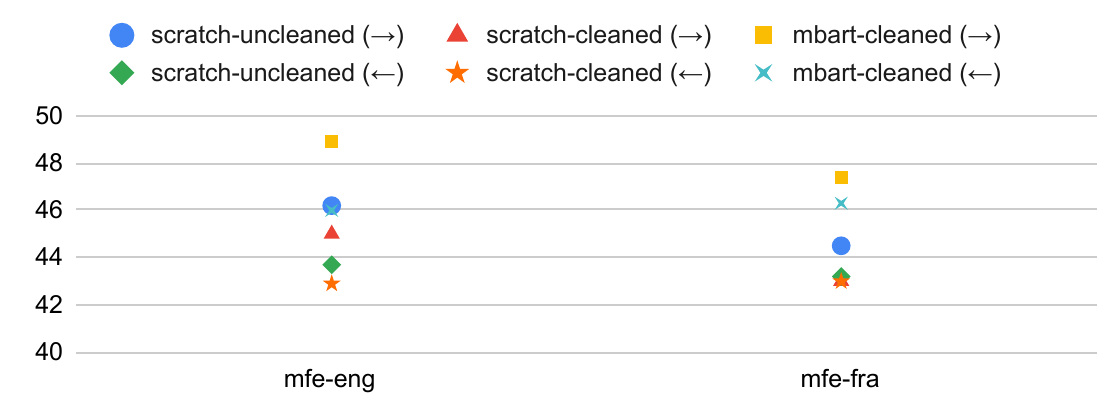}
    \caption{Results on the KreolMorisienMT test sets using the models trained on \textbf{all} (top) and \textbf{public} (bottom) data.}
    \label{fig:kreolmorisien-mt-results}
\end{figure}

\subsection{Lego-MT and KreolMorisienMT Results}

\begin{figure*}[t]
    \centering
    \includegraphics[width=\textwidth]{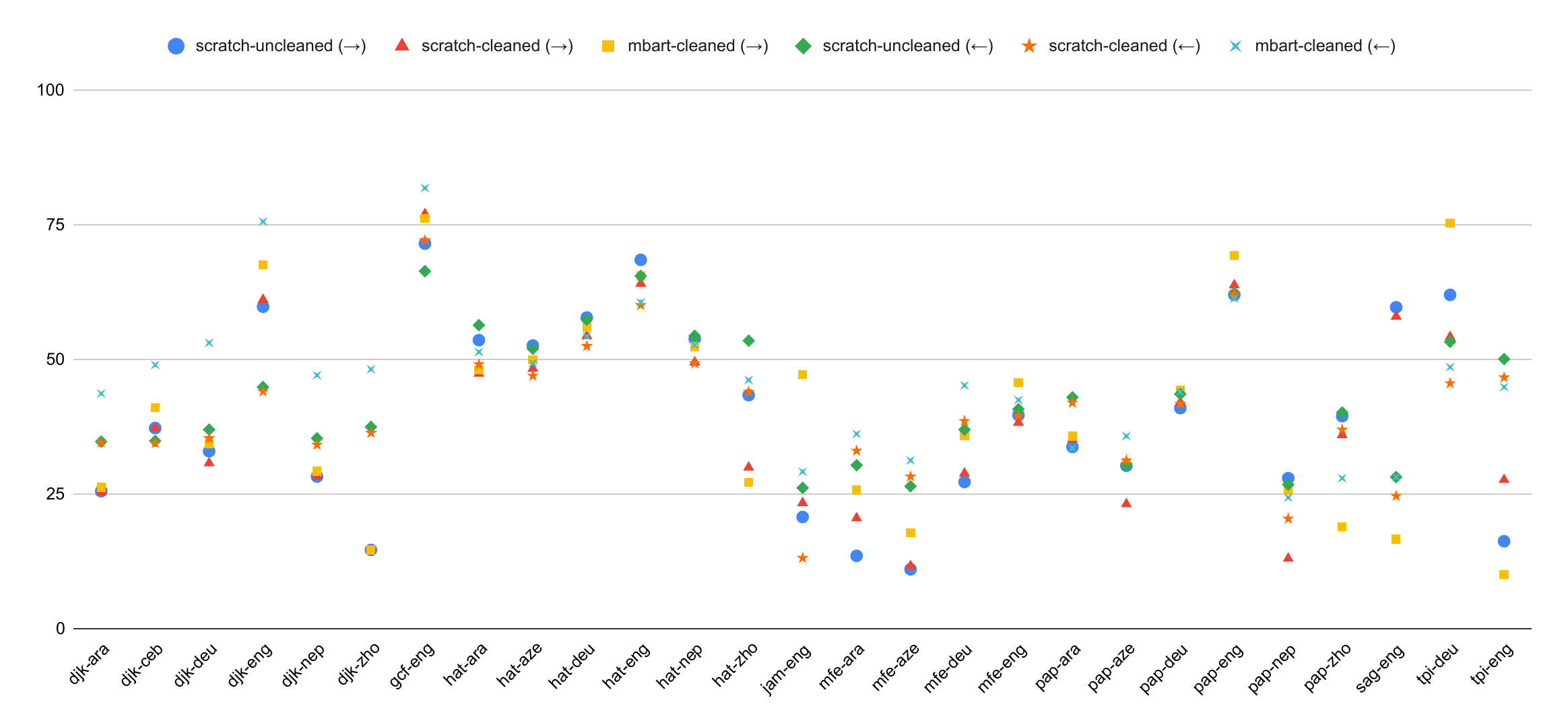}
    \includegraphics[width=\textwidth]{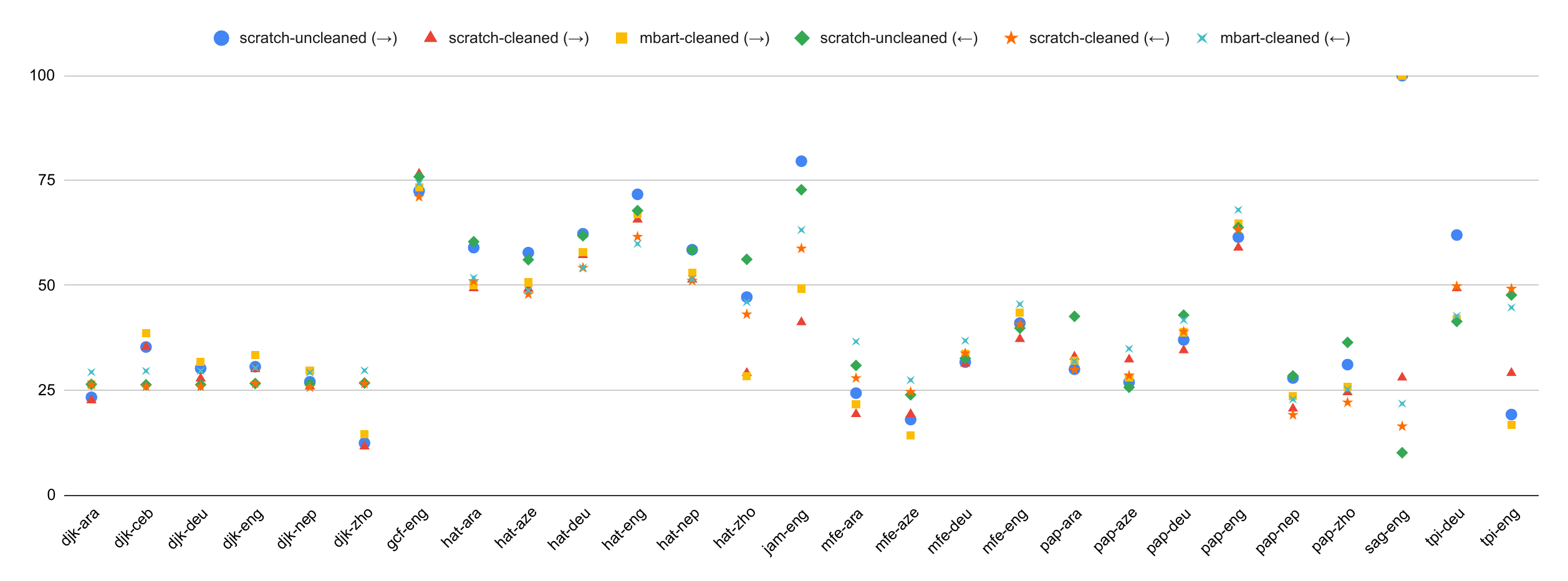}
    \caption{Results on the Lego-MT test sets using the models trained on \textbf{all} (top) and \textbf{public} (bottom) data.}
    \label{fig:lego-mt-results}
\end{figure*}

Figure~\ref{fig:kreolmorisien-mt-results}  shows results of our models on the KreolMorisienMT \cite{dabre2022kreolmorisienmt} test sets, and Figure~\ref{fig:lego-mt-results} shows results of the same models on the Lego-MT \cite{yuan2023lego} test sets. 
\citet{yuan2023lego} did not report results on their own test splits; hence we do not compare our models' performance directly with theirs. 
Once again, our fine-tuned models give the best performance.

\begin{table*}
    \centering
    \small
    \begin{tabular}{
        >{\ttfamily}{l}
        *{8}{r}
    }
        \toprule
               & \textbf{CreoleVal} & \textbf{JHU} & \textbf{Lego-MT} & \textbf{FLORES} & \textbf{AfricaNLP} & \textbf{NLLB} & \textbf{Ours} & \textbf{Ours} \\
        \midrule
        {\rmfamily \textbf{Public?}} & {\xmark} & {\xmark} & {\cmark} & {\cmark} & {\xmark} & {\cmark} & {\cmark} & {\xmark} {\cmark} \\
        \midrule
        acf & 7864 & 15989 & {-} & {-} & {-} & {-} & 4406 & 23916  \\
        aoa & {-} & {-} & {-} & {-} & {-} & {-} & 198 & 198  \\
        bah & {-} & {-} & {-} & {-} & {-} & {-} & 327 & 327  \\
        brc & {-} & {-} & {-} & {-} & {-} & {-} & 222 & 222  \\
        bzj & 14911 & 23406 & {-} & {-} & {-} & {-} & 229 & 31002  \\
        bzk & {-} & {-} & {-} & {-} & {-} & {-} & 391 & 391  \\
        cab & {-} & 20879 & {-} & {-} & {-} & {-} & {-} & 20879  \\
        cri & {-} & {-} & {-} & {-} & {-} & {-} & 306 & 306  \\
        crs & 222613 & 5055 & {-} & {-} & {-} & {-} & 3186 & 225875  \\
        dcr & {-} & {-} & {-} & {-} & {-} & {-} & 189 & 189  \\
        djk & 45361 & 23748 & 7868 & {-} & {-} & {-} & 15266 & 68833  \\
        fab & {-} & {-} & {-} & {-} & {-} & {-} & 204 & 204  \\
        fng & {-} & {-} & {-} & {-} & {-} & {-} & 160 & 160  \\
        fpe & {-} & {-} & {-} & {-} & {-} & {-} & 259 & 259  \\
        gcf & {-} & {-} & 96 & {-} & {-} & {-} & 6467 & 6467  \\
        gcr & {-} & {-} & {-} & {-} & {-} & {-} & 1433 & 1433  \\
        gpe & {-} & {-} & {-} & {-} & {-} & {-} & 223 & 223  \\
        gul & 7870 & 7990 & {-} & {-} & {-} & {-} & 266 & 8831  \\
        gyn & {-} & {-} & {-} & {-} & {-} & {-} & 258 & 258  \\
        hat & 210593 & 72354 & 477048 & 2006 & 179435 & 4256455 & 5715227 & 6023034  \\
        icr & 7799 & 15702 & {-} & {-} & {-} & {-} & 317 & 16774  \\
        jam & 7988 & 25206 & 26 & {-} & 5118 & {-} & 434 & 28713  \\
        kea & {-} & {-} & {-} & 2009 & {-} & 129449 & 132931 & 132931  \\
        kri & 50438 & 23740 & {-} & {-} & {-} & {-} & 185 & 66736  \\
        ktu & 7886 & 5055 & {-} & {-} & {-} & {-} & 175 & 10737  \\
        lou & {-} & {-} & {-} & {-} & {-} & {-} & 1860 & 1860  \\
        mart1259 & {-} & {-} & {-} & {-} & {-} & {-} & 5153 & 5153  \\
        mfe & 191909 & 23625 & 399 & {-} & {-} & {-} & 25633 & 233320  \\
        mue & {-} & {-} & {-} & {-} & {-} & {-} & 147 & 147  \\
        pap & 397354 & 5018 & 269 & 2009 & {-} & 4898029 & 4968965 & 5363394  \\
        pcm & 31128 & 15905 & {-} & {-} & {-} & {-} & 8084 & 47455  \\
        pov & {-} & {-} & {-} & {-} & {-} & {-} & 480 & 480  \\
        pre & {-} & {-} & {-} & {-} & {-} & {-} & 243 & 243  \\
        rcf & {-} & {-} & {-} & {-} & {-} & {-} & 285 & 285  \\
        sag & 262334 & 16952 & 9 & 2009 & {-} & 235749 & 260560 & 535310  \\
        srm & 42303 & 23531 & {-} & {-} & {-} & {-} & 440 & 59053  \\
        srn & 583830 & 24569 & {-} & {-} & {-} & {-} & 6620 & 615010  \\
        svc & {-} & {-} & {-} & {-} & {-} & {-} & 321 & 321  \\
        tpi & 398341 & 81595 & 70 & 2009 & {-} & 424626 & 451758 & 925648  \\
        trf & {-} & {-} & {-} & {-} & {-} & {-} & 1691 & 1691  \\
        \bottomrule
    \end{tabular}
    \caption{Size of total bitext data collected for Creole languages to date, compared with our full combined bitext sets. Bitext size is measured as the number of unique Creole language segments paired with a translation in any target language
    }
    \label{tab:data_amt}
\end{table*}

\begin{table*}
\centering
\small
\begin{tabular}{l*{8}{r}}
\toprule
\textbf{Lang}     & \textbf{Bible}    & \textbf{Educational} & \textbf{Legal}    & \textbf{Narrative} & \textbf{News}    & \textbf{Religious} & \textbf{Wikipedia} & \textbf{Other/Mix}  \\
\midrule
acf      & 15989  & 4406        & 0      & 0         & 0     & 0         & 0         & 33778     \\
aoa      & 0      & 198         & 0      & 0         & 0     & 0         & 0         & 0         \\
bah      & 0      & 327         & 0      & 0         & 0     & 0         & 0         & 0         \\
brc      & 0      & 222         & 0      & 0         & 0     & 0         & 0         & 0         \\
bzj      & 54933  & 229         & 0      & 0         & 0     & 0         & 0         & 10213     \\
bzk      & 0      & 391         & 0      & 0         & 0     & 0         & 0         & 0         \\
cab      & 20879  & 0           & 0      & 0         & 0     & 0         & 0         & 47471     \\
cri      & 0      & 306         & 0      & 0         & 0     & 0         & 0         & 0         \\
crs      & 5055   & 273         & 443948 & 15719     & 4141  & 0         & 0         & 279331    \\
dcr      & 0      & 189         & 0      & 0         & 0     & 0         & 0         & 0         \\
djk      & 23815  & 7398        & 0      & 0         & 0     & 0         & 0         & 66491     \\
fab      & 0      & 204         & 0      & 0         & 0     & 0         & 0         & 0         \\
fng      & 0      & 160         & 0      & 0         & 0     & 0         & 0         & 0         \\
fpe      & 0      & 259         & 0      & 0         & 0     & 0         & 0         & 0         \\
gcf      & 1559   & 304         & 0      & 4446      & 0     & 0         & 0         & 158       \\
gcr      & 879    & 159         & 0      & 2388      & 0     & 0         & 15141     & 0         \\
gpe      & 0      & 223         & 0      & 0         & 0     & 0         & 12425     & 0         \\
gul      & 7990   & 262         & 0      & 0         & 0     & 0         & 0         & 579       \\
gyn      & 0      & 258         & 0      & 0         & 0     & 0         & 0         & 0         \\
hat      & 71958  & 9359        & 0      & 4         & 0     & 115583    & 5452      & 5828317   \\
icr      & 15702  & 317         & 0      & 0         & 0     & 0         & 0         & 755       \\
jam      & 18420  & 233         & 0      & 0         & 0     & 0         & 4588      & 29647     \\
kea      & 0      & 6108        & 0      & 0         & 229   & 0         & 0         & 132314    \\
kri      & 16113  & 185         & 0      & 0         & 0     & 0         & 0         & 99454     \\
ktu      & 5055   & 175         & 0      & 0         & 0     & 0         & 0         & 62584     \\
lou      & 0      & 668         & 0      & 1192      & 0     & 0         & 0         & 0         \\
mart1259 & 0      & 283         & 0      & 0         & 0     & 0         & 0         & 4870      \\
mfe      & 23624  & 258         & 0      & 274       & 0     & 0         & 0         & 277331    \\
mue      & 0      & 147         & 0      & 0         & 0     & 0         & 0         & 0         \\
pap      & 5018   & 2996        & 0      & 0         & 0     & 65573     & 92        & 7430445   \\
pcm      & 15905  & 253         & 0      & 0         & 0     & 0         & 0         & 31297     \\
pov      & 0      & 480         & 0      & 0         & 0     & 0         & 0         & 0         \\
pre      & 0      & 243         & 0      & 0         & 0     & 0         & 0         & 0         \\
rcf      & 0      & 285         & 0      & 0         & 83659 & 0         & 0         & 63975     \\
sag      & 16952  & 192         & 0      & 0         & 0     & 0         & 0         & 518166    \\
srm      & 23531  & 440         & 0      & 0         & 0     & 0         & 0         & 57013     \\
srn      & 24567  & 6607        & 0      & 0         & 0     & 0         & 4600      & 766364    \\
svc      & 0      & 321         & 0      & 0         & 0     & 0         & 0         & 0         \\
tpi      & 81178  & 62          & 0      & 0         & 0     & 25018     & 7         & 819383    \\
trf      & 0      & 174         & 0      & 0         & 0     & 0         & 0         & 1517      \\
wes      & 0      & 223         & 0      & 0         & 0     & 0         & 0         & 0         \\
\midrule
\textbf{Total}    & 449122 & 45777       & 443948 & 24023     & 88029 & 206174    & 42305     & 16561453\\
\bottomrule
\end{tabular}
\caption{Amount of sentences for each language in each genre.}
\label{tab:app_genre}
\end{table*}

\begin{table*}
    \centering
    \small
    \begin{tabular}{
        l
        rrrrr
        rrr
    }
        \toprule
        \multicolumn{5}{c}{\textbf{Bitext}} & \multicolumn{3}{c}{\textbf{Monolingual}}\\
        \cmidrule(lr){1-5}
        \cmidrule(lr){6-8}
        & 
        \multicolumn{2}{c}{\textit{Web}} & \multicolumn{2}{c}{\textit{PDF}} & & & \\
        \cmidrule(lr){2-3}
        \cmidrule(lr){4-5}
        Lang     & \textit{Prev. pub.} & \textit{aligned} & \textit{articles} & \textit{aligned} & \textit{other} & \textit{Prev. pub.} & \textit{Web} & \textit{PDF} \\
        \midrule
        acf      & 19510               & 0                & 0                 & 4406             & 0              & 30257               & 0            & 0            \\
        aoa      & 0                   & 198              & 0                 & 0                & 0              & 0                   & 0            & 0            \\
        bah      & 0                   & 327              & 0                 & 0                & 0              & 0                   & 0            & 0            \\
        brc      & 0                   & 222              & 0                 & 0                & 0              & 0                   & 0            & 0            \\
        bzj      & 30773               & 229              & 0                 & 0                & 0              & 0                   & 34373        & 0            \\
        bzk      & 0                   & 391              & 0                 & 0                & 0              & 0                   & 0            & 0            \\
        cab      & 20879               & 0                & 0                 & 0                & 0              & 47471               & 0            & 0            \\
        cri      & 0                   & 306              & 0                 & 0                & 0              & 0                   & 0            & 0            \\
        crs      & 222690              & 273              & 0                 & 0                & 2912           & 61696               & 445177       & 15719        \\
        dcr      & 0                   & 189              & 0                 & 0                & 0              & 0                   & 0            & 0            \\
        djk      & 61435               & 7398             & 0                 & 0                & 0              & 28871               & 0            & 0            \\
        fab      & 0                   & 204              & 0                 & 0                & 0              & 0                   & 0            & 0            \\
        fng      & 0                   & 160              & 0                 & 0                & 0              & 0                   & 0            & 0            \\
        fpe      & 0                   & 259              & 0                 & 0                & 0              & 0                   & 0            & 0            \\
        gcf      & 64                  & 4844             & 1559              & 0                & 0              & 0                   & 0            & 0            \\
        gcr      & 0                   & 159              & 880               & 0                & 394            & 0                   & 15140        & 1994         \\
        gpe      & 0                   & 223              & 0                 & 0                & 0              & 0                   & 12425        & 0            \\
        gul      & 8569                & 262              & 0                 & 0                & 0              & 0                   & 0            & 0            \\
        gyn      & 0                   & 258              & 0                 & 0                & 0              & 0                   & 0            & 0            \\
        hat      & 5900275             & 165              & 122594            & 0                & 0              & 0                   & 7639         & 0            \\
        icr      & 16457               & 317              & 0                 & 0                & 0              & 0                   & 0            & 0            \\
        jam      & 28311               & 233              & 169               & 0                & 0              & 19756               & 4419         & 0            \\
        kea      & 131454              & 484              & 860               & 0                & 133            & 0                   & 229          & 5491         \\
        kri      & 66551               & 185              & 0                 & 0                & 0              & 49016               & 0            & 0            \\
        ktu      & 10562               & 175              & 0                 & 0                & 0              & 57077               & 0            & 0            \\
        lou      & 0                   & 440              & 0                 & 228              & 1192           & 0                   & 0            & 0            \\
        mart1259 & 0                   & 231              & 0                 & 4870             & 52             & 0                   & 0            & 0            \\
        mfe      & 232788              & 258              & 0                 & 274              & 0              & 68167               & 0            & 0            \\
        mue      & 0                   & 147              & 0                 & 0                & 0              & 0                   & 0            & 0            \\
        pap      & 5294750             & 146              & 65665             & 2833             & 0              & 2140730             & 0            & 0            \\
        pcm      & 47202               & 253              & 0                 & 0                & 0              & 0                   & 0            & 0            \\
        pov      & 0                   & 480              & 0                 & 0                & 0              & 0                   & 0            & 0            \\
        pre      & 0                   & 243              & 0                 & 0                & 0              & 0                   & 0            & 0            \\
        rcf      & 0                   & 285              & 0                 & 0                & 0              & 60098               & 83659        & 3877         \\
        sag      & 535118              & 192              & 0                 & 0                & 0              & 0                   & 0            & 0            \\
        srm      & 58613               & 440              & 0                 & 0                & 0              & 21931               & 0            & 0            \\
        srn      & 608397              & 606              & 4                 & 6003             & 0              & 182532              & 4596         & 0            \\
        svc      & 0                   & 321              & 0                 & 0                & 0              & 0                   & 0            & 0            \\
        tpi      & 900560              & 63               & 25025             & 0                & 0              & 0                   & 0            & 0            \\
        trf      & 1517                & 174              & 0                 & 0                & 0              & 0                   & 0            & 0            \\
        wes      & 0                   & 223              & 0                 & 0                & 0              & 0                   & 0            & 0            \\
        \midrule
        Total    & 14196475                     & 21963               & 216756               & 18614               & 4683              & 2767602                    & 607657    & 27081\\
        \bottomrule
    \end{tabular}
    \caption{Number of segments gathered from each source type/extraction method for each language}
    \label{tab:app_dtype}
\end{table*}

\begin{table*}[!ht]
    \centering
    \small
    \begin{tabular}{lllll}
        \textbf{ISO} & \textbf{Name 1} & \textbf{Name 2} & \textbf{Name 3} & \textbf{Name 4} \\ \hline
        gcf & French Antillean & Guadeloupean Creole & Martinican & - \\ \hline
        gcr & French Guianese & Kriyòl Gwiyannen & Kriyòl Lagwiyann & Gwiyannen \\ \hline
        djk & Ndyuka & Eastern Maroon Creole & Aukan & Nengee \\ \hline
        kmv & Karip\'{u}na & Amap\'{a} Creole & Uaç\'{a} Creole & - \\ \hline
        bzk & Miskito Coast Creole & Nicaraguan Creole English & - & - \\ \hline
        pcm & Naija & Nigerian Pidgin & - & - \\ \hline
        kea & Cape Verdean & Kabuverdianu & - & - \\ \hline
        mfe & Mauritian Creole & Morisyen & - & - \\ \hline
        crs & Seychellois & Seselwa & Kreol Sesel & - \\ 
    \end{tabular}
    \caption{Alternate names for languages with more than one common name. The ISO-639 code and all names used for searches in our data collection process are listed.}
    \label{tab:lang_aliases}
\end{table*}

\clearpage
\onecolumn

\section{Attributions} \label{sec:attribution}

We provide exact attributions for all our data sources here. 
We list them with string identifiers that we used to distinguish them in our own data organization. 
See our repository \github{} for data downloading instructions. 

\subsection{Resources for Bitexts}

\textbf{APiCS} - Atlas of Pidgin and Creole Language Structures \cite{apics}

\begin{itemize}

\item
\textbf{Languages}: \textit{dcr, icr, bzj, pov, fng, trf, rcf, fpe, kri, pap, gcf, gpe, tpi, crs, ktu, pre, fab, bah, srm, gyn, djk, brc, sag, aoa, pcm, svc, mart1259, bzk, cri, gcr, kea, wes, hat, lou, srn, jam, mue, mfe, gul}
\item
\textbf{Links}: \url{https://apics-online.info/}
\end{itemize}

\noindent \textbf{AfricaNLP-2023} - Aligned sentenced pairs from \cite{robinson2023african}

\begin{itemize}

\item
\textbf{Languages}: \textit{hat, jam}
\item
\textbf{Links}: \url{https://openreview.net/forum?id=YKUv4sSOom}
\end{itemize}

\noindent \textbf{bible\_uedin} - bible-uedin-v1, Parallel corpus created from translations of the Bible \cite{christodouloupoulos2015massively}

\begin{itemize}

\item
\textbf{Languages}: \textit{djk}
\item
\textbf{Links}: \url{https://opus.nlpl.eu/bible-uedin/corpus/version/bible-uedin}
\end{itemize}

\noindent \textbf{Bidze2019} - Seychelles Government Budget For the Fiscal Year 2019, Office of the President of The Republic of Seychelles

\begin{itemize}

\item
\textbf{Languages}: \textit{crs}
\item
\textbf{Links}: \url{https://www.statehouse.gov.sc/downloads?page=2}
\end{itemize}

\noindent \textbf{Bidze2021} - Seychelles Government Budget For the Fiscal Year 2021, Office of the President of The Republic of Seychelles

\begin{itemize}

\item
\textbf{Languages}: \textit{crs}
\item
\textbf{Links}: \url{https://www.statehouse.gov.sc/downloads?page=1}
\end{itemize}

\noindent \textbf{boston-food-forest} - 
Boston Food Forest Coalition flyers translations

\begin{itemize}

\item
\textbf{Languages}: \textit{kea}
\item
\textbf{Links}: \url{https://www.bostonfoodforest.org/languages}
\end{itemize}

\noindent \textbf{CJCLDS} - Online library of The Church of Jesus Christ of Latter-day Saints

\begin{itemize}

\item
\textbf{Languages}: \textit{pap, hat, tpi}
\item
\textbf{Links}: \url{https://www.churchofjesuschrist.org/study?lang=pap} (Link to full LDC dataset available on our repository: \github{}.)
\end{itemize}

\noindent \textbf{CREOLORAL} - Martinican and Guadeloupean oral corpus with annotations \cite{glaude2012creoloral}

\begin{itemize}

\item
\textbf{Languages}: \textit{gcf}
\item
\textbf{Links}: \url{https://cocoon.huma-num.fr/exist/crdo/search2.xql?lang=fr&language=http%3A%2F%2Flexvo.org%2Fid%2Fiso639-3%2Fgcf}
\end{itemize}

%\textbf{Confiant-Dictionary}
\noindent \textbf{Confiant-Dictionary} - Dictionnaire Créole Martiniquais - Français, Raphaël Confiant
\cite{confiant2007dictionnaire}

\begin{itemize}

\item
\textbf{Languages}: \textit{mart1259}
\item
\textbf{Links}: \url{https://www.potomitan.info/dictionnaire/}
\end{itemize}

\noindent \textbf{CreoleVal} \cite{lent2023creoleval}

\begin{itemize}

\item
\textbf{Languages}: \textit{djk, kri, icr, pap, hat, bzj, sag, ktu, acf, srn, pcm, tpi, jam, crs, mfe, gul, srm}
\item
\textbf{Links}: \url{https://arxiv.org/abs/2310.19567}
\end{itemize}

\noindent \textbf{dicoNengee} - Dictionnaire Nengee - Français - English 

\begin{itemize}

\item
\textbf{Languages}: \textit{djk}
\item
\textbf{Links}: \url{https://corporan.huma-num.fr/Lexiques/dicoNengee.html}
\end{itemize}

\noindent \textbf{FLORES-200} \cite{team2022language}

\begin{itemize}

\item
\textbf{Languages}: \textit{kea, pap, hat, sag, tpi}
\item
\textbf{Links}: \url{https://github.com/facebookresearch/flores/blob/main/flores200}
\end{itemize}

\noindent \textbf{folklore} - Excerpts from \textit{Le folklore de l'Ile-Maurice (texte créole et traduction française)} \cite{baissac1888folk}

\begin{itemize}

\item
\textbf{Languages}: \textit{mfe}
\item
\textbf{Links}: \url{https://archive.org/details/lefolkloredelile00bais/page/98/mode/2up}
\end{itemize}

\noindent \textbf{fortier} - Excerpts from \textit{Louisiana Folk-tales: In French Dialect and English Translation} \cite{fortier1895louisiana}

\begin{itemize}

\item
\textbf{Languages}: \textit{lou}
\item
\textbf{Links}: \url{https://archive.org/details/b24865424/page/n11/mode/2up}
\end{itemize}

\noindent \textbf{GoiloText} - Papiamentu Textbook, E.R. Goilo

\begin{itemize}

\item
\textbf{Languages}: \textit{pap}
\item
\textbf{Links}: \url{https://archive.org/details/PapiamentuTextbook/mode/2up}
\end{itemize}

\noindent \textbf{JHU} - The Johns Hopkins University Bible Corpus \cite{mccarthy2020johns}

\begin{itemize}

\item
\textbf{Languages}: \textit{djk, kri, icr, pap, hat, bzj, sag, cab, ktu, acf, srn, pcm, tpi, jam, crs, mfe, gul, srm}
\item
\textbf{Links}: \url{https://aclanthology.org/2020.lrec-1.352/}
\end{itemize}

\noindent \textbf{kapes} - Corrections of the "Certificat d'aptitude au professorat de l'enseignement du second degré" (CAPES) exam for Martinican and Guadeloupean Creole

\begin{itemize}

\item
\textbf{Languages}: \textit{gcf, mart1259}
\item
\textbf{Links}: \url{https://kapeskreyol.potomitan.info/}
\end{itemize}

\noindent \textbf{KreolMorisienMT} \cite{dabre2022kreolmorisienmt}

\begin{itemize}

\item
\textbf{Languages}: \textit{mfe}
\item
\textbf{Links}: \url{https://aclanthology.org/2022.findings-aacl.3.pdf}
\end{itemize}

\noindent \textbf{LAFANDMT} \cite{adelani-etal-2022-masakhaner}

\begin{itemize}

\item
\textbf{Languages}: \textit{pcm}
\item
\textbf{Links}: \url{https://github.com/masakhane-io/lafand-mt}
\end{itemize}

\noindent \textbf{LegoMT} \cite{yuan-etal-2023-lego}

\begin{itemize}

\item
\textbf{Languages}: \textit{djk, pap, gcf, hat, sag, tpi, jam, mfe}
\item
\textbf{Links}: \url{https://aclanthology.org/2023.findings-acl.731/}
\end{itemize}

\noindent \textbf{mindelo} - Online dictionary

\begin{itemize}

\item
\textbf{Languages}: \textit{kea}
\item
\textbf{Links}: \url{http://www.mindelo.info/\_dico.php}
\end{itemize}

\noindent \textbf{MIT-Haiti}, MIT-Haiti Initiative \cite{lent2023creoleval}

\begin{itemize}

\item
\textbf{Languages}: \textit{hat}
\item
\textbf{Links}: \url{https://haiti.mit.edu/hat/resous/}
\end{itemize}

\noindent \textbf{MiBelNouvel} - Translation of the Gospel of John in Guadeloupean Creole

\begin{itemize}

\item
\textbf{Languages}: \textit{gcf}
\item
\textbf{Links}: \url{https://mibelnouvel.wordpress.com/}
\end{itemize}

\noindent \textbf{MultiCCAligned} \cite{tiedemann-2012-parallel,elkishky_ccaligned_2020}

\begin{itemize}

\item
\textbf{Languages}: \textit{hat}
\item
\textbf{Links}: \url{https://opus.nlpl.eu/MultiCCAligned.php}
\end{itemize}

\noindent \textbf{NLLB} NLLB-v1 \cite{team2022language, schwenk-etal-2021-ccmatrix}

\begin{itemize}

\item
\textbf{Languages}: \textit{hat, tpi, sag}
\item
\textbf{Links}: \url{https://opus.nlpl.eu/NLLB/corpus/version/NLLB}, \url{https://huggingface.co/datasets/allenai/nllb}
\end{itemize}

\noindent \textbf{PwovebKreyol} - Proverbes \& expressions créoles

\begin{itemize}

\item
\textbf{Languages}: \textit{gcf}
\item
\textbf{Links}: \url{http://pwoveb.kreyol.free.fr/proverbes.php}
\end{itemize}

\noindent \textbf{QCRI} - \cite{abdelali-etal-2014-amara}

\begin{itemize}

\item
\textbf{Languages}: \textit{pap}
\item
\textbf{Links}: \url{https://opus.nlpl.eu/QED/corpus/version/QED}
\end{itemize}

\noindent \textbf{QED} - \cite{lamm-etal-2021-qed}

\begin{itemize}

\item
\textbf{Languages}: \textit{mfe, hat, pap, sag}
\item
\textbf{Links}: \url{https://aclanthology.org/2021.tacl-1.48/}
\end{itemize}

\noindent \textbf{quentin} - Excerpts from \textit{Introduction à l'histoire de Cayenne ; suivie d'un Recueil de contes, fables et chansons en créole} \cite{saintquentin1872introduction}

\begin{itemize}

\item
\textbf{Languages}: \textit{gcr}
\item
\textbf{Links}: \url{https://gallica.bnf.fr/ark:/12148/bpt6k82939m.r=creole%20guyanais%20quentin?rk=21459;2}
\end{itemize}

\noindent \textbf{SIL-Suriname} - Languages of Suriname, SIL

\begin{itemize}

\item
\textbf{Languages}: \textit{djk, srm}
\item
\textbf{Links}: \url{https://suriname-languages.sil.org/Aukan/Aukan.html}, \url{https://suriname-languages.sil.org/Saramaccan/Saramaccan.html}
\end{itemize}

\noindent \textbf{Saint\_Lucia\_Ministry\_of\_Ed} - Kwéyòl
Dictionary, Ministry of Education
Government of Saint Lucia

\begin{itemize}

\item
\textbf{Languages}: \textit{acf}
\item
\textbf{Links}: \url{http://www.saintluciancreole.dbfrank.net/dictionary/KweyolDictionary.pdf}
\end{itemize}

\noindent \textbf{TEC-English} - Trinidad English Creole to English Dataset, University of the West Indies \cite{smith2022trinidad}

\begin{itemize}

\item
\textbf{Languages}: \textit{trf}
\item
\textbf{Links}: \url{https://data.mendeley.com/datasets/n4259kw9y7/1}
\end{itemize}

\noindent \textbf{TED2020} - TED and TED-X transcripts \cite{reimers-2020-multilingual-sentence-bert}

\begin{itemize}

\item
\textbf{Languages}: \textit{hat}
\item
\textbf{Links}: \url{https://opus.nlpl.eu/TED2020.php}
\end{itemize}

\noindent \textbf{Tatoeba} - Tatoeba database

\begin{itemize}

\item
\textbf{Languages}: \textit{pap, gcf, sag, srn, tpi, jam, mfe, hat}
\item
\textbf{Links}: \url{https://tatoeba.org/en/downloads}
\end{itemize}

\noindent \textbf{TiLiv} - Ti Liv Kreyol, A learner’s guide to Louisiana creole \cite{guillory2020tiliv}

\begin{itemize}

\item
\textbf{Languages}: \textit{lou}
\item
\textbf{Links}: \url{https://dn790005.ca.archive.org/0/items/ti-liv-kreyol-second-edition/Ti%20Liv%20Kreyol%20Second%20Edition.pdf}
\end{itemize}

\noindent \textbf{Ubuntu} - Ubuntu Translations

\begin{itemize}

\item
\textbf{Languages}: \textit{hat, pap}
\item
\textbf{Links}: 
\url{https://object.pouta.csc.fi/OPUS-Ubuntu/v14.10/moses/en-ht.txt.zip}, \url{https://object.pouta.csc.fi/OPUS-Ubuntu/v14.10/moses/en_AU-ht.txt.zip}, \url{https://object.pouta.csc.fi/OPUS-Ubuntu/v14.10/moses/en_CA-ht.txt.zip}, \url{https://object.pouta.csc.fi/OPUS-Ubuntu/v14.10/moses/en_GB-ht.txt.zip}, \url{https://object.pouta.csc.fi/OPUS-Ubuntu/v14.10/moses/en-pap.txt.zip}, \url{https://object.pouta.csc.fi/OPUS-Ubuntu/v14.10/moses/en_AU-pap.txt.zip}, \url{https://object.pouta.csc.fi/OPUS-Ubuntu/v14.10/moses/en_CA-pap.txt.zip}, \url{https://object.pouta.csc.fi/OPUS-Ubuntu/v14.10/moses/en_GB-pap.txt.zip}

\end{itemize}

\noindent \textbf{Wikimedia}

\begin{itemize}

\item
\textbf{Languages}: \textit{pap, hat, tpi, jam, srn, gcr}
\item
\textbf{Links}: \url{https://en.wikipedia.org/}
\end{itemize}

\noindent \textbf{Wikipedia}

\begin{itemize}

\item
\textbf{Languages}: \textit{gcf}
\item
\textbf{Links}: \url{https://fr.wikipedia.org/wiki/Cr%C3%A9ole\_martiniquais}
\end{itemize}

\noindent \textbf{Wortubuku} - Wortubuku fu Sranan Tongo, Sranan Tongo - English Dictionary \cite{loninghing2007wortubuku}

\begin{itemize}

\item
\textbf{Languages}: \textit{srn}
\item
\textbf{Links}: \url{https://www.sil.org/resources/archives/1538}
\end{itemize}

\noindent \textbf{XLEnt} \cite{el-kishky-etal-2021-xlent}

\begin{itemize}

\item
\textbf{Languages}: \textit{hat}
\item
\textbf{Links}: \url{https://aclanthology.org/2021.emnlp-main.814/}
\end{itemize}

\noindent \textbf{YouVersion–Bible} - Life.Church online Bible

\begin{itemize}

\item
\textbf{Languages}: \textit{gcr}
\item
\textbf{Links}: \url{https://www.bible.com/bible/2963/JHN.INTRO1.GCR07}
\end{itemize}

\subsection{Resources for Monolingual Corpora}

\noindent \textbf{Anacao} - Articles from the \textit{A Nação} newspaper

\begin{itemize}

\item
\textbf{Languages}: \textit{kea}
\item
\textbf{Links}: \url{https://www.anacao.cv/,}
\end{itemize}

\noindent \textbf{atipa} - Excerpts from the book \textit{Atipa} \cite{parépou1987atipa}

\begin{itemize}

\item
\textbf{Languages}: \textit{gcr}
\item
\textbf{Links}: \url{https://www.google.com.ng/books/edition/\_/F7bA4J4D6T4C?hl=en&kptab=overview}
\end{itemize}

\noindent \textbf{Belizean} - Life.Church online Bible

\begin{itemize}

\item
\textbf{Languages}: \textit{bzj}
\item
\textbf{Links}: \url{https://www.bible.com/bible/409/MAT.1.BZJ}
\end{itemize}

\noindent \textbf{Creolica} - Online corpus of Seychellois Créole

\begin{itemize}

\item
\textbf{Languages}: \textit{rcf, crs}
\item
\textbf{Links}: \url{https://creolica.net/Corpus-de-creole-seychellois}, \url{https://creolica.net/Corpus-de-creole-reunionnais}
\end{itemize}

\noindent \textbf{Fonologia} - Extracted interviews from \citet{rodrigues2017fonologia}

\begin{itemize}

\item
\textbf{Languages}: \textit{kea}
\item
\textbf{Links}: \url{https://core.ac.uk/download/pdf/33531609.pdf}
\end{itemize}

\noindent \textbf{graelo} - Wikipedia dumps

\begin{itemize}

\item
\textbf{Languages}: \textit{gcr}
\item
\textbf{Links}: \url{https://huggingface.co/datasets/graelo/wikipedia}
\end{itemize}

\noindent \textbf{JamPatoisNLI} - \cite{armstrong2022jampatoisnli}

\begin{itemize}

\item
\textbf{Languages}: \textit{jam}
\item
\textbf{Links}: \url{https://arxiv.org/abs/2212.03419}
\end{itemize}

\noindent \textbf{KreolMorisienMT} - \cite{dabre-sukhoo-2022-kreolmorisienmt}

\begin{itemize}

\item
\textbf{Languages}: \textit{mfe}
\item
\textbf{Links}: \url{https://aclanthology.org/2022.findings-aacl.3.pdf}
\end{itemize}

\noindent \textbf{MADLAD-400} - \cite{kudugunta2023madlad}

\begin{itemize}

\item
\textbf{Languages}: \textit{mfe, pap, crs, kri, srm, jam, srn, djk, ktu, acf, rcf, cab, bzj}
\item
\textbf{Links}: \url{https://arxiv.org/abs/2309.04662}
\end{itemize}

\noindent \textbf{MIT-Haiti} - Learning Resources from the MIT-Haiti initiative \cite{lent2023creoleval}

\begin{itemize}

\item
\textbf{Languages}: \textit{hat}
\item
\textbf{Links}: \url{https://haiti.mit.edu/hat/resous/}
\end{itemize}

\noindent \textbf{National} - Official minutes of the Sittings of the House, The National Assembly of Seychelles

\begin{itemize}

\item
\textbf{Languages}: \textit{crs}
\item
\textbf{Links}: \url{https://www.nationalassembly.sc/verbatim}
\end{itemize}

\noindent \textbf{Seychelles} - Articles from the \textit{NATION}  newspaper, National Information Services Agency (NISA) 

\begin{itemize}

\item
\textbf{Languages}: \textit{crs}
\item
\textbf{Links}: \url{http://nation.sc/}
\end{itemize}

\noindent \textbf{temoignages} - Articles from the \textit{Témoignages} newpaper

\begin{itemize}

\item
\textbf{Languages}: \textit{rcf}
\item
\textbf{Links}: \url{https://www.temoignages.re/chroniques/ote/}
\end{itemize}

\noindent \textbf{Wikidumps}

\begin{itemize}

\item
\textbf{Languages}: \textit{jam, gpe, gcr, srn}
\item
\textbf{Links}: \url{https://huggingface.co/datasets/graelo/wikipedia/viewer}
\end{itemize}

\noindent \textbf{Wikimedia}

\begin{itemize}

\item
\textbf{Languages}: \textit{srn}
\item
\textbf{Links}: \url{https://archive.org/details/srnwiki-20180101}
\end{itemize}

\end{document}